\documentclass[reprint,
aps,pre,twoside
]{revtex4-2}

\usepackage{graphicx}% Include figure files
\usepackage[justification=raggedright, singlelinecheck=false]{caption}
\usepackage[justification=centering, singlelinecheck=false]{subcaption}
\usepackage{dcolumn}% Align table columns on decimal point
\usepackage{bm}% bold math
\usepackage{diagbox}
\usepackage{amsthm}
\usepackage{amssymb}
\usepackage{amsmath}
\usepackage{algorithm,algcompatible}
\usepackage{algpseudocode}
\usepackage[dvipsnames]{xcolor}
\usepackage{algorithmicx}

\newtheorem{theorem}{Theorem}
\newtheorem{lemma}{Lemma}
\newtheorem{definition}{Definition}

\algnewcommand\Input{\item[\textbf{Input:}]}
\algnewcommand\Output{\item[\textbf{Output:}]}

\makeatletter
\AtBeginDocument{
  \@twosidetrue
  
  \def\ps@headings{
    \def\@evenhead{\small\thepage\hfill\textit{Physical Review E}, In Press, July 2026}%
    
    \def\@oddhead{\small\textit{Physical Review E}, In Press, July 2026\hfill\thepage}%
  }
  \ps@headings
}
\makeatother

\begin{document}

\preprint{APS/123-QED}

\title{A Bayesian Approach for the Network Reconstruction of Interdependent Critical Infrastructure Systems from Cascading Failures}% Force line breaks with \\
\thanks{MirSaleh Bahavarnia and Yu Wang have equally contributed to the manuscript. The corresponding author is Hiba Baroud.}

\author{MirSaleh Bahavarnia}
\email{mirsaleh.bahavarnia@vanderbilt.edu}
\affiliation{Department of Civil and Environmental Engineering, Vanderbilt University, Nashville, TN, USA.}

\author{Yu Wang}
\email{yuwang@uoregon.edu}
\affiliation{School of Computer and Data Sciences at the University of Oregon, Eugene, OR, USA.}

\author{Jin-Zhu Yu}
\email{jinzhu.yu@uta.edu}
\affiliation{Department of Civil Engineering, University of Texas at Arlington, Arlington, TX, USA.}

\author{Hiba Baroud}
\email{hiba.baroud@vanderbilt.edu}
\affiliation{Department of Civil and Environmental Engineering, Vanderbilt University, Nashville, TN, USA.}

\begin{abstract}
Analyzing the behavior of complex interdependent networks requires complete information about the network topology and the interdependent links across networks. For many applications, such as interdependent critical infrastructure (ICI) networks, understanding network interdependencies is crucial to anticipate cascading failures and mitigate the risk from disruptions. However, complete network data is often unavailable due to security concerns, and some important interdependent links are only revealed in the aftermath of a disruption. This study formulates and solves a network reconstruction problem to uncover uncertain network interdependencies during disruptions. We propose a scalable nonparametric Bayesian approach to reconstruct the topology of ICI networks from (observed) cascading failures. Metropolis-Hastings (M-H) algorithm coupled with the infrastructure-dependent proposal (IP) is employed to increase the efficiency of sampling possible graphs. Numerical results of reconstructing a synthetic system of ICI networks demonstrate that the proposed approach outperforms existing methods in both accuracy and computational time.
\end{abstract}

\maketitle

\section{Introduction} \label{sec: introduction}

Networks offer a powerful tool for modeling and analyzing diverse systems with complex interactions, such as ecological, social, biological, and technological systems \cite{fath2007ecological,clauset2008hierarchical,brugere2018network,sharma2019communication}. Examples include modeling interdependent critical infrastructure (ICI) systems (e.g., coupled water and power systems) as multi-layer networks to analyze their vulnerability and resilience to disruptive events \cite{ten2008vulnerability}. By definition, \textit{interdependency} refers to bidirectional dependencies between infrastructure networks through which individual networks interact, influencing overall system operations \cite{wang2021generating,rinaldi2001identifying}. There are primarily four categories of interdependencies \cite{rinaldi2001identifying,ouyang2014review}: \textit{(i)} physical, \textit{(ii)} cyber, \textit{(iii)} geographic, and \textit{(iv)} logical. This study examines physical interdependencies. Additionally, \textit{multi-layer} networks in this context refer to different layers inside each critical infrastructure network, including supply, transmission, and demand layers. Ideally, performing such network-level assessments requires data from real infrastructure networks with complete information on the network topology, flow, and interdependencies. However, data on the topology of real-world infrastructure networks is often \textit{unavailable} either due to security concerns or decentralized operations of different infrastructure sectors \cite{hamilton2017representation}. As a result, infrastructure network performance has been evaluated using \textit{model-driven} techniques relying on assumptions of the existence of interdependencies. Recent research advances have explored synthetic infrastructure data generation to overcome real network data challenges \cite{ahmad2020synthetic, zhai2021power,wang2021generating}. In this study, we propose a \textit{data-driven} technique founded in network reconstruction methods to infer the structure of complex interdependent infrastructure networks based on cascading failure data containing independent cascading failure scenarios (See Definition 1 in Section \ref{sec: framework} for mathematical details). A cascading failure occurs when an initial failure in an interconnected system triggers successive dependent failures through positive feedback loops. This process typically continues until no further failures occur or the system stabilizes. A failure of a small fraction of nodes in one network can initiate an iterative cascade of failures in several interdependent networks \cite{rinaldi2001identifying}. This work constitutes the first attempt at developing network reconstruction methods for solving data and uncertainty challenges in complex infrastructure systems.

Network reconstruction from observations of the dynamics on the target network is a fundamental yet challenging \textit{inverse problem} \cite{braunstein2019network,peixoto2019network,gray2020bayesian}. The research advancements centering around this important problem can be categorized into (\textit{i}) capturing network structure using partial information \cite{zhao2017link,vajdi2018identification}, and (\textit{ii}) anticipating dynamic changes \cite{zhang2018reconstructing, wilinski2020scalable}. While applications such as social, biological, and ecological networks have obvious dynamic patterns, the structure of critical infrastructure networks has been assumed to be \textit{static} and \textit{deterministic} unless the network is subject to uncertain disruptions. Network reconstruction has been applied to an individual critical infrastructure network scenario, such as road networks \cite{dey2019road}; however, reconstructing interdependencies of multiple infrastructure sectors is still lacking. To achieve more representative network interdependencies, we aim to propose a \textit{data-driven} network reconstruction approach to infer the structure of ICI networks from (observed) cascading failures.

\subsection{Background and Related Work}\label{sec: background}

Network reconstruction aims to \textit{infer} the network topology from direct or indirect data that are inherently connected to the underlying network, such as observations of dynamic processes on the network \cite{clauset2008hierarchical,brugere2018network}. Instances include the diffusion of news among social media \cite{louni2014diffusion}, the propagation of cascading failures in infrastructure networks \cite{little2002controlling}, and the spreading of influence among political parties \cite{parmelee2011politics}, among others. Although a single sequence of cascading failures may not provide sufficient information on the underlying network, combining sufficiently rich sequences of cascading failures can enable robust network reconstruction to obtain valuable insights into the functionality of the underlying network \cite{pajevic2009efficient}. Data-driven network reconstruction approaches can be divided into: (\textit{i}) graph embedding-based approaches \cite{kipf2016semi,kipf2016variational,grover2016node2vec,hamilton2017inductive}, (\textit{ii}) optimization-based approaches \cite{madni2013systems,shen2014reconstructing}, and (\textit{iii}) the Bayesian approaches \cite{peixoto2019network,braunstein2019network,gray2020bayesian}. The first two approaches require information that is typically unavailable for critical infrastructure, whereas Bayesian approaches are commonly used to uncover the complete network structure from partial observations while accounting for \textit{uncertainty}. Bayesian approaches for network reconstruction are either (\textit{i}) parametric or (\textit{ii}) nonparametric. Parametric approaches require the network type to be known, whereas many real-world systems are comprised of a diverse set of interconnected networks that do not fall into a well-defined network type. Nonparametric approaches employ a probabilistic approach via a sufficient number of samples of possible topologies to identify the \textit{most probable} network. Prior studies equip Bayesian approaches with the epidemic spreading models and the information diffusion cascade models to infer structure from network dynamics \cite{peixoto2019network,gray2020bayesian}. This study develops a nonparametric Bayesian approach to infer the discrete posterior distribution for the topology of the target infrastructure network from observed cascading failures.

\subsection{Paper Contributions}\label{sec: contributions}

In addition to the novel application of network reconstruction approaches in ICIs, this work advances the methodology in two ways. First, we devise an infrastructure-dependent proposal (IP) that incorporates topological and physical constraints of critical infrastructure and significantly reduces the sampling space in the application of the Markov Chain Monte Carlo (MCMC) algorithm for network reconstruction. Using the block structure of ICI networks, we leverage the hierarchical stochastic block model (HSBM) to calculate the prior probability of the potential graphs. Moreover, the convergence of the MCMC algorithm is ensured and numerically demonstrated. The second methodological contribution lies in the design of three computational optimization techniques to improve the efficiency of the MCMC algorithm in reconstructing large-scale networks where the number of possible network topologies increases exponentially over the number of nodes. The performance of these three techniques is numerically demonstrated via experiments on reconstructing synthetic ICI networks.

The remainder of this paper is structured as follows: Section \ref{sec: framework} introduces the general Bayesian model for reconstructing the network topology based on cascading failures data. The MCMC algorithm with the infrastructure-dependent proposal (IP) to infer the Bayesian model is described in Section \ref{sec: MCMC}. In Section \ref{sec: computationopt}, we present three computational optimization techniques to improve the computational efficiency of the proposed network reconstruction approach. Extensive numerical experiments for validating the proposed approach are shown in Section \ref{sec: results}, followed by concluding remarks in Section \ref{sec: conclusions}.

\section{Bayesian Approach for Network Reconstruction}\label{sec: framework}

\begin{definition}[Network Reconstruction] Given a set of time series data $\bm{C}$ about the binary state of nodes that are generated from a cascading failure process on a network with unknown adjacency matrix $\bm{A}$, the network structure is inferred by finding the adjacency matrix $\bm{A}$ with the highest conditional probability $P(\bm{A}|\bm{C})$. The set of cascading failure data $\bm{C}$ contains independent cascading failure scenarios (sequences) $\bm{c}^i, i \in \{1,\dots,C\}$ starting from time $1$ (discrete-time setup) where $\bm{c}^i_{t,j} = 1$  ($\bm{c}^i_{t-1,j} = 0$) indicates node $j$ fails at time $t$ in the $i$-th cascading failure scenario. The failure of node $j$ (probabilistically) occurs when at least one of its adjacent nodes fails, and this failure propagates (with a probability) along edges to node $j$. We leverage the susceptible-infected (SI) epidemic model to simulate the cascading failure across infrastructure sectors. We also assume the independence between any two distinct cascading failure scenarios. Additionally, we assume that the unknown adjacency matrix $\bm{A}$ is time-invariant. Similarly, we assume that the probability of failure propagating from a node to its adjacent node does not depend on time, and we ensure that the failure propagation is Markovian.
\end{definition}

In the Bayesian approach for network reconstruction, we estimate the discrete posterior distribution of the network topology given the observations, i.e., a set of initial node failure scenarios and the following cascading failure sequences. Specifically, conditioned on the cascading failure data $\bm{C}$, we estimate the discrete posterior distribution $P(\bm{A}|\bm{C})$ for the adjacency matrix $\bm{A}$ of the underlying network via Bayes' rule \cite{peixoto2019network,gray2020bayesian} as
\begin{align} \label{eq:bayesianupdate}
    & P(\bm{A}|\bm{C}) = {P(\bm{C}|\bm{A})P(\bm{A})}/{P(\bm{C})} \propto P(\bm{C}|\bm{A})P(\bm{A}),
\end{align}
where $P(\bm{C}|\bm{A})$ is the likelihood that the cascading failures $\bm{C}$ occur in a network with topology $\bm{A}$, $P(\bm{A})$ encodes the prior information on the network topology, and $P(\bm{C})$ is the normalization constant which represents the total evidence for the cascading failure data $\bm{C}$ \cite{peixoto2019network}. Details on calculating the graph prior $P(\bm{A})$ and the likelihood $P(\bm{C}|\bm{A})$ are presented in the following two sections.

\subsection{Hierarchical Stochastic Block Model (HSBM)}\label{sec: HSBM}

Calculating the $P(\bm{A})$ requires realistic possible samples of the target network to be reconstructed. Various types of networks having various prior probabilities $P(\bm{A})$ can be generated by various graph models. The adopted graph model should represent realistic behaviors of the target network. Generative graph models for real networks, such as the classical random, scale-free, and small-world networks, cannot represent real-world ICIs as nodes in infrastructure networks are grouped by blocks and hierarchy \cite{ouyang2014review}. Moreover, random graphs will generate unrealistic networks that do not comply with the physical constraints and properties of real infrastructure networks. As such, a suitable model for characterizing ICI networks would be the HSBM \cite{yu2020modeling}.

The HSBM is built upon the Stochastic Block Model (SBM), which divides different nodes into different blocks based on the node memberships. The HSBM adds hierarchical structures in each block according to node functionality levels. The probability of an edge between two nodes depends on the blocks to which the nodes belong and the hierarchical levels at which the nodes are positioned. Considering a multi-layer network with a set of blocks $\mathcal{M}$ and the number of nodes in each block is $n_{M}, M\in\mathcal{M}$, each single block $M \in \mathcal{M}$ has a hierarchical structure $\mathcal{L}$ where $n_{M}$ nodes are further divided into $|\mathcal{L}|$ levels. Considering the set of nodes and edges in the multi-layer network as $\mathcal{V}^{\mathcal{M}} = \bigcup\limits_{M\in\mathcal{M}}{\mathcal{V}^M}$ and $\mathcal{E}^{\mathcal{M}} = \bigcup\limits_{M\in\mathcal{M}}{\mathcal{E}^M}$, and denoting the block and the level labels of the node $k$ as $m_k\in\mathcal{M}$ and $l_k \in \mathcal{L}$, the probability of an edge $(k,j)$ is an independent Bernoulli random variable $p_{kj}$ conditioned on the block and the level labels $m_k, m_j, l_k, l_j$ of nodes $k$, $j$, respectively. Then, the prior probability conditioned on the block and the layer assignment is calculated as
\begin{subequations}
\begin{align}
    & P(\bm{A}|\mathcal{M}, \mathcal{L}) = \prod\limits_{\forall k, j\in \mathcal{V}^{\mathcal{M}},k \ne j}{p_{kj}^{\bm{A}_{kj}}(1 - p_{kj})^{1 - \bm{A}_{kj}}}\\
    & p_{kj} = g(m_{k}, m_{j}, l_{k}, l_{j})
\end{align}    
\end{subequations}where $g()$ is an HSBM function calculating the edge probability based on the block and the layer assignments of nodes. Conditioning the graph prior probability on the HSBM model, we introduce prior knowledge on the network topology $P(\bm{A}|\mathcal{M}, \mathcal{L})$ from the function $g()$. In ICIs, $\mathcal{M}$ corresponds to individual infrastructure sectors (e.g., power grid and gas distribution network) and $\mathcal{L}$ represents different types of facilities with different functionalities (e.g., power supply and distribution nodes).

\subsection{Cascading Failure Model}\label{sec: cascading failure}

After employing the HSBM to model the prior information of the network topology, we use the cascading failure model to calculate the likelihood in \eqref{eq:bayesianupdate}. Denoting the adjacency matrix of the network to be reconstructed as $\bm{A}^{\ast}$, and the prior knowledge about the network topology is updated with the likelihood $P(\bm{C}|\bm{A})$ towards the target topology $\bm{A}^{\ast}$ determining the diffusion of the cascading failure encoded in $\bm{C}$. Disasters cause initial failures of infrastructure components, which trigger flow redistribution of resources and energy within and across the infrastructure networks \cite{zhang2019robustness}, causing overload and damage to additional components, leading to cascading failures until the system reaches a new steady-state with no newly-failed components. We leverage the susceptible-infected (SI) epidemic model to simulate the cascading failure across infrastructure sectors, wherein the facility failure is modeled by the node \textit{infections} and the flow redistribution is modeled by the probabilistic propagation of the infection in the network.

Due to the independence between any two distinct cascading failure scenarios, the likelihood of the cascading failure data $\bm{C}$ encodes the failure propagation across all cascading failure scenarios $i \in \{1,\dots,C\}$, over all time-steps for the entire disruption duration $\{1,\dots,T(\bm{c}^i)\}$, and among all nodes in the multi-layer network $j\in\mathcal{V}^{\mathcal{M}}$. Such a likelihood is given as
\begin{align} \label{eq:likelihood}
    & P(\bm{C}|\bm{A}) = \prod\limits^{C}_{i=1}{\prod\limits_{t = 1}^{T(\bm{c}^i)}{\prod\limits_{j=1}^{|\mathcal{V}^{\mathcal{M}}|}{P(\bm{c}^i_{t +1,j}|\bm{c}^i_{t})}}}
\end{align}where $T(\bm{c}^i)$ is the number of time-steps that the cascading failure scenario $\bm{c}^i$ lasts and $P(\bm{c}^i_{t+1, j}|\bm{c}^i_t)$ is the probability of the status of operation of node $j$ at time-step $t + 1$, which is calculated by incorporating operable node status as
{\small \begin{align*}
& P(\bm{c}^i_{t+1,j}|\bm{c}^i_{t}) = \nonumber\\& {\Bigg( {1 - {\prod _{k \in  {{\mathcal V}^{\mathcal M}}\backslash \{j\} }}\left[ {1 - {q_{kj}}{\bm{A}_{kj}}\bm{c}_{t,k}^i (1 - {\rm{ }}\bm{c}_{t - 1,k}^i)} \right]} \Bigg) ^{\bm{c}_{t + 1,j}^i(1 - \bm{c}_{t,j}^i)}}\nonumber\\
& \times {\Bigg( { {\prod _{k \in {{\mathcal V}^{\mathcal M}}\backslash \{j\} }}\left[ {1 - {q_{kj}}{\bm{A}_{kj}}\bm{c}_{t,k}^i(1 - {\rm{ }}\bm{c}_{t - 1,k}^i)} \right]} \Bigg) ^{(1-\bm{c}_{t + 1,j}^i)(1 - \bm{c}_{t,j}^i)}}
\end{align*}}where $q_{kj}$ denotes the probability of failure propagating from node $k$ to node $j$. The term $\bm{c}^i_{t, k}(1 - \bm{c}^i_{t-1, k})$ ensures that the failure propagation is \textit{Markovian}, i.e., the failure of nodes at the current time-step is only impacted by nodes that failed at the last time-step, matching the failure process of infrastructure networks, such as the progressive collapse of structures and gradual outages of the power stations \cite{adam2018research}.

\section{MCMC for network reconstruction} \label{sec: MCMC}

By generating possible network topologies with the graph prior model from the HSBM and incorporating the cascading failure model, \eqref{eq:bayesianupdate} becomes
{\small \begin{align} \label{eq:bayesianfull}
    & P(\bm{A}|\bm{C}, \mathcal{M}, \mathcal{L}) \propto  \underbrace{P(\bm{C}|\bm{A}, \mathcal{M}, \mathcal{L})}_{\text{likelihood}}\underbrace{P(\bm{A} | \mathcal{M}, \mathcal{L}) P(\mathcal{M}, \mathcal{L})}_{\text{prior probability}}.
\end{align}}Since the cascading failure scenarios $\bm{C}$ are independent of the blocks and the layers assignments, the likelihood $P(\bm{C}|\bm{A}, \mathcal{M}, \mathcal{L})$ reduces to $P(\bm{C}|\bm{A})$. Then, since $P(\mathcal{M}, \mathcal{L})$ does not contain $\bm{A}$, \eqref{eq:bayesianfull} reduces to
\begin{align} \label{eq:finalform}
    & P(\bm{A}|\bm{C}, \mathcal{M}, \mathcal{L})  \propto P(\bm{C}|\bm{A}) P(\bm{A}|\mathcal{M}, \mathcal{L}).
\end{align}

We clarify that while our approach utilizes a Hierarchical Stochastic Block Model (HSBM) and an SI likelihood---both of which involve parameter estimation---we characterize the method as nonparametric in the sense that it avoids assuming a rigid, predefined network topology, allowing the latent structure to be inferred directly from the data.

\subsection{Metropolis-Hastings (M-H) Algorithm} \label{sec: metropolis}

Since \eqref{eq:finalform} generally does not admit a closed-form solution, simulation techniques are usually leveraged to generate samples of the discrete posterior distributions of possible network topologies. MCMC methods such as the Metropolis-Hastings (M-H) algorithm and Gibbs sampling algorithm are commonly used to perform Bayesian inference via sampling \cite{cheung2009bayesian,yu2021hierarchical}. We choose the M-H algorithm over the Gibbs sampling algorithm as the latter requires an analytical solution to the conditional distributions of each model parameter. At each iteration, a new multi-layer network $\mathcal{M}'$ with topology $\bm{A}'$ is proposed based on the current network $\mathcal{M}$ with topology $\bm{A}$ via the proposal distribution $Q(\bm{A}'|\bm{A})$. The algorithm probabilistically accepts the new proposal with the following acceptance ratio:
\begin{align} \label{eq:mcmcupdate}
    \gamma = \min \bigg\{ 1, \frac{P(\bm{C}|\bm{A}')P(\bm{A}'|\mathcal{M},\mathcal{L})}{P(\bm{C}|\bm{A})P(\bm{A}|\mathcal{M},\mathcal{L})}\frac{Q(\bm{A}|\bm{A}')}{Q(\bm{A}'|\bm{A})} \bigg \}.
\end{align}

The performance of the algorithm relies on the choice of the $Q(\bm{A}'|\bm{A})$. The random-walk graph proposal is a common choice where a random node pair is selected either by \textit{adding} new edges or \textit{removing} the existing edges between them. However, this approach can generate \textit{invalid} or \textit{unrealistic} proposals for infrastructure networks. For instance, proposals of possible water-power networks, a likely edge is from pumping stations to storage tanks, representing water extraction from nearby rivers at pumping stations and its transfer to storage tanks to be stored for future use. Nevertheless, random graph generation may add edges going from storage tanks to pumping stations, which is unrealistic. We thus devise an \textit{infrastructure-dependent proposal} (IP) method that imposes additional topology constraints on the proposed networks.

\subsection{Infrastructure-dependent Proposal (IP)}\label{sec: proposal}

The infrastructure-dependent proposal (IP) ensures that only networks with a topology conforming to ICIs are admissible. This is achieved by first analyzing the standard topology of ICIs, from which we abstract the topological constraints for designing the IP. Each network in ICIs has a hierarchical structure with three levels (i.e., $|\mathcal{L}| = 3$) corresponding to \textit{supply}, \textit{transmission}, and \textit{demand} facilities. Given a set of infrastructure networks $\mathcal{M}$ where supply, transmission, and demand nodes are denoted as $\textit{s}$, $\textit{t}$, and $\textit{d}$, and a set of interdependent links $\mathcal{I}$ across the networks (each interdependent link $I \in \mathcal{I}$ is associated with an ordered pair of infrastructure networks), we can express the topological constraints as follows:
\begin{enumerate}
     \item $\forall M\in \mathcal{M}, \forall k\in \mathcal{V}_{\text{s}}^M, \exists j\in \mathcal{V}_{\text{d}}^M, k \mathop {\rightarrow}\limits^{\text{path}} j$. \label{constraint1}
     \item $\forall M\in \mathcal{M}, \forall k \in \mathcal{V}_{\text{d}}^M, \exists j\in \mathcal{V}_{\text{s}}^M, j\mathop {\rightarrow}\limits^{\text{path}} k$.\label{constraint2}
     \item $\forall M\in \mathcal{M}, \forall k\in \mathcal{V}_{\text{t}}^M, \exists j\in \mathcal{V}_{\text{d}}^M, k\mathop {\rightarrow}\limits^{\text{edge}} j$.\label{constraint3}
     \item $\forall M\in \mathcal{M}, \forall k\in \mathcal{V}_{\text{t}}^M, \exists j\in \mathcal{V}_{\text{s}}^M, j\mathop {\rightarrow}\limits^{\text{edge}} k$.\label{constraint4}
     \item $\forall I\in \mathcal{I}, \forall k \in \mathcal{V}_{\text{d}}^I, \exists j\in \mathcal{V}_{\text{s}}^I, k\mathop {\rightarrow}\limits^{\text{edge}} j$.\label{constraint5}
     \item $\forall I\in \mathcal{I}, \forall k \in \mathcal{V}_{\text{s}}^I, \exists j\in \mathcal{V}_{\text{d}}^I, j\mathop {\rightarrow}\limits^{\text{edge}} k$.\label{constraint6}
     \item $\forall M\in \mathcal{M}, \forall (k, j)\in \mathcal{E}^M, k\in \mathcal{V}_1, j\in\mathcal{V}_2, (\mathcal{V}_1, \mathcal{V}_2) \in \{(\mathcal{V}^M_{\text{s}}, \mathcal{V}^M_{\text{t}}), (\mathcal{V}^M_{\text{t}}, \mathcal{V}^M_{\text{d}}), (\mathcal{V}^M_{\text{s}}, \mathcal{V}^M_{\text{d}})\}$.\label{constraint7}
     \item $\forall I\in \mathcal{I}, \forall (k, j)\in \mathcal{E}^I, k\in \mathcal{V}^I_d, j\in\mathcal{V}^I_s$.\label{constraint8}
     \item $\forall M\in \mathcal{M}$, no cycles in $M$.\label{constraint9}
     \item $\forall I\in \mathcal{I}$, no cycles in $I$.\label{constraint10}
\end{enumerate}

Customarily, resources are generated at the supply nodes and transported via the transmission nodes to the demand nodes, where the resources are further distributed to the residents. Thus, in every infrastructure network $M\in\mathcal{M}$, every supply node $k \in \mathcal{V}^M_{\text{s}}$ is connected by a \textit{path} to at least one demand node $j \in \mathcal{V}^M_{\text{d}}$ and vice versa, corresponding to constraints \ref{constraint1}--\ref{constraint2}. Every transmission node $k \in \mathcal{V}^M_{\text{t}}$ is connected by an \textit{edge} to at least one demand node $j \in \mathcal{V}^M_{\text{d}}$, corresponding to the constraint \ref{constraint3}. Also, every transmission node $k \in \mathcal{V}^M_{\text{t}}$ is connected by an edge from at least one supply node $j \in \mathcal{V}^M_{\text{d}}$, corresponding to the constraint \ref{constraint4}. Similarly, for every interdependent link $I \in \mathcal{I}$, we have the same constraints for connectivity from the demand nodes to the supply nodes, according to constraints \ref{constraint5}--\ref{constraint6}. The resources move from the supply nodes (high-level) to the demand nodes (low-level) with no resources flowing back (either directly or indirectly via the transmission nodes), which results in only forward edges from the high-level nodes to the low-level nodes according to constraint \ref{constraint7}, which automatically leads to no cycles in each infrastructure network as shown by constraint \ref{constraint9}. For every interdependent link, the directions of the edges are from the demand nodes to the supply nodes according to constraint \ref{constraint8}, which automatically leads to no cycles in each interdependent link, as shown by constraint \ref{constraint10}. The rationale behind the existence of the transmission nodes is driven by the physical nature of infrastructure networks. In most spatial network models, transmission nodes are employed to represent the aggregation points. While constraint \ref{constraint7} allows for direct supply-to-demand edges, the intermediate transmission nodes are necessary for two main reasons: \textit{(i)} dimensionality reduction (avoiding highly dense network connections via a limited set of transmission nodes), and \textit{(ii)} economies of scale (enabling unit cost minimization via bulk handling at aggregated transmission nodes). From a sampling perspective, multi-hop paths (i.e., having sequentially connected transmission nodes) significantly increase the diameter of the graph. This often slows down the convergence of MCMC-based sampling schemes because the proposal must now account for a chain of dependencies rather than a simple bipartite or tripartite structure. Therefore, we have not included $(\mathcal{V}_t^M,\mathcal{V}_t^M)$ in constraint \ref{constraint7}.

The M-H algorithm with the IP and considerations of the $10$ constraints is outlined in Alg. $1$. First, a basic adjacency matrix $\bm{A}^0$ is generated following lines $1$--$6$ to initiate the algorithm. While we construct $\bm{A}^0$ by connecting failed nodes in sequential time-steps, any method that generates graphs with $P(\bm{C}|\bm{A}) >0$ can be used. In lines $10$--$11$, we randomly select a node pair $(k,j)$ from the subset of the feasible set $\mathcal{S}=\{(\mathcal{V}^{M}_{\text{s}}, \mathcal{V}^{M}_{\text{t}}), (\mathcal{V}^{M}_{\text{s}}, \mathcal{V}^{M}_{\text{d}}), (\mathcal{V}^{M}_{\text{t}}, \mathcal{V}^{M}_{\text{d}}), (\mathcal{V}^{I}_{\text{d}}, \mathcal{V}^{I}_{\text{s}})\}, \forall M\in\mathcal{M}, \forall I\in\mathcal{I}\}$, which guarantees the satisfaction of constraints \ref{constraint7}--\ref{constraint10}. We then \textit{add} or \textit{remove} the edge between that node pair $(k,j)$ and generate a new candidate network following lines $12$--$16$. The candidate network is \textit{accepted} as a sample of the discrete posterior distribution of $\mathcal{A}$, if the addition or removal of this edge does not violate constraints \ref{constraint1}--\ref{constraint6}. Otherwise, we \textit{reject} this candidate, propose another node pair, and check the feasibility of the proposed topology, as shown in lines $17$--$25$.

The advantage of this IP over the traditional (original) one lies in the improvement in the sampling efficiency. By imposing the additional topological constraints, the invalid topology proposals are excluded, thereby increasing the accuracy of sampling. Also, such exclusion of the invalid topology proposals shrinks the space of the candidate topology (See Appendix \ref{app: reducespace}), which significantly decreases the mixing time of the M-H algorithm. We also prove the convergence of the Markov chain formed by the IP (See Appendix \ref{app:converge}).

{\small \par\noindent\rule{0.483\textwidth}{0.5pt}
{\normalsize \textbf{Alg. 1:} M-H algorithm with the IP}
\par\noindent\rule{0.483\textwidth}{0.4pt}
\begin{algorithmic}[1] 
  \Input{$\mathcal{M}$, $\mathcal{I}$, $\mathcal{V}^M_{\text{s}}, \mathcal{V}^M_{\text{t}}, \mathcal{V}^M_{\text{d}}$, $\bm{C}$, $\text{Iter}_{\max}$, $\mathcal{S}$}
  \Output{$\mathcal{A}$}
  \For{cascading failure scenario $i \in \{1,\dots, C\}$}
    \For{$t \in \{1,\dots, T(\bm{c}^i) - 1\}$}
        \State Add an edge between every node failing at $t$ and\\
            \hspace{2.85em} every node failing at $t+1$
    \EndFor
\EndFor\Comment{\textit{Initialize the adjacency matrix $\bm{A}^0$}}
\State $\text{Iter} \leftarrow 0$
\While{$\text{Iter} \le \text{Iter}_{\max}$}\Comment{\textit{Start M-H sampling}}
  \State $\bm{A} \leftarrow \bm{A}^{\text{Iter}}$
  \State Randomly pick a node pair $(k, j)$, $k\in\mathcal{V}_1$, $j\in$ $ \mathcal{V}_2$ and\\\hspace{1.22em} the set $(\mathcal{V}_1, \mathcal{V}_2) \in \mathcal{S}$\Comment{\textit{Impose constraints} \ref{constraint7}--\ref{constraint10}}
  \If{$\bm{A}_{kj} = 0$}
    \State $\bm{A}_{kj} \leftarrow 1$
  \Else
    \State $\bm{A}_{kj} \leftarrow 0$
  \EndIf\Comment{\textit{Add or remove the edge}}
   \If{$\bm{A}$ causes no violation of constraints \ref{constraint1}--\ref{constraint6}}
   \State Calculate the acceptance ratio $\gamma$ by \eqref{eq:mcmcupdate}
   \If{$\gamma \ge \tilde{p} \sim U(0,1)$}
   \State Accept $\bm{A}$ as a sample of the posterior\\
          \hspace{4.36em} distribution of $\mathcal{A}$
   \State $\text{Iter} \leftarrow \text{Iter} + 1$
   \State $\bm{A}^{\text{Iter}} \leftarrow \bm{A}$
   \EndIf
   \EndIf\Comment{\textit{Accept or reject the sample}}
\EndWhile
\State \textbf{return} $\mathcal{A}$
\end{algorithmic}
\par\noindent\rule{0.483\textwidth}{0.5pt}}

\section{Sampling Efficiency Improvements} \label{sec: computationopt}

Although the IP significantly shrinks the space of the candidate network topology and thus improves the sampling efficiency, the computational expense can still be very high due to the computation of proposing and updating the topology of the network at each iteration. At each iteration, lines $8$--$26$ in Alg. $1$ take $\mathcal{O}(1)$ operations to randomly choose node pairs from the feasible set and add or remove the associated edges. Evaluating the likelihood $P(\bm{C}|\bm{A})$ in calculating the acceptance ratio $\gamma$ takes $\mathcal{O}(\sum_{i=1}^{C}{T(\bm{c}^i)|\mathcal{V}^\mathcal{M}|^2})$ time to double-loop over all nodes in $\mathcal{V}^\mathcal{M}$ at time-step $t$ to calculate the failure probability of every node at $t+1$ and iterate over all time-steps in $\{1,\dots,T(\bm{c}^i) - 1\}$ through all cascading failure scenarios $\bm{c}^i, i\in\{1,\dots,C\}$. Validating whether the generated network $\bm{A}$ violates constraints \ref{constraint1}--\ref{constraint6} requires $\mathcal{O}(|\mathcal{V}^\mathcal{M}|(|\mathcal{V}^\mathcal{M}| + |\mathcal{E}^\mathcal{M}|) + |\mathcal{V}^\mathcal{I}|(|\mathcal{V}^\mathcal{I}| + |\mathcal{E}^\mathcal{I}|))$---$\mathcal{V}^\mathcal{I} := \bigcup_{I \in \mathcal{I}} \mathcal{V}^{I}$ and $\mathcal{E}^{\mathcal{I}} = \bigcup_{I \in \mathcal{I}} \mathcal{E}^{I}$---as we check the existence of paths by tree-search algorithms such as Depth First Search (DFS) or Breadth First Search (BFS) on every node in every infrastructure network and every interdependent link. Each iteration in Alg. $1$ takes 
{\footnotesize \begin{align*}
    & {\mathcal O}(\underbrace {\sum\limits_{i = 1}^C {T({\bm{c}^i})|{{\mathcal V}^{\mathcal M}}} {|^2}}_{{\text{Likelihood calculation}}} + \underbrace {|\mathcal{V}^\mathcal{M}|(|\mathcal{V}^\mathcal{M}| + |\mathcal{E}^\mathcal{M}|) + |\mathcal{V}^\mathcal{I}|(|\mathcal{V}^\mathcal{I}| + |\mathcal{E}^\mathcal{I}|)}_{{\text{Graph validation}}})
\end{align*}}which is nonlinear due to $|\mathcal{V}^\mathcal{M}|^2$ in the likelihood calculation and the $|\mathcal{V}^{\mathcal{M}}|^2 + |\mathcal{V}^{\mathcal{I}}|^2$ in the graph validation. Since the interdependent infrastructure network is a sparse and tripartite graph (s-t-d), we devise the following three optimization techniques to speed up the computation:

\subsection{Likelihood Calculation}\label{subsec: likelihood calculation}

The quadratic term $|\mathcal{V}^\mathcal{M}|^2$ is due to evaluating the failure probability of each node $j$ by considering failure propagation from all other nodes, $1 - \prod\limits_{k\in \mathcal{V}^\mathcal{M}\backslash \{j\}}(1 - q_{kj}\bm{A}_{kj}\bm{c}^i_{t, k}(1 - \bm{c}^i_{t-1, k}))$. This computation can be simplified by only considering the failed neighborhoods of node $j$, which is $1 - {\prod _{k \in  {{\mathcal{N}}^{\mathcal{M}}}(j) }}(1 - {q_{kj}}{\bm{c}}_{t,k}^i(1 - {\bm{c}}_{t - 1,k}^i))$. Instead of double-looping over all nodes, we traverse the \textit{edge list} and integrate the failure propagation probability over edges with failed heading nodes and the same tailing nodes. Since the whole process requires iterating over the adjacency list (an array of linked lists describing the neighbors of a particular node), the time complexity of the likelihood calculation reduces to $\mathcal{O}(\sum_{i=1}^{C}{T(\bm{c}^i)(|\mathcal{V}^\mathcal{M}| + |\mathcal{E}^\mathcal{M}|)})$.

\subsection{Graph Validation}\label{subsec: graphvalidation}

The second-order term $|\mathcal{V}^\mathcal{M}|(|\mathcal{V}^\mathcal{M}| + |\mathcal{E}^\mathcal{M}|) + |\mathcal{V}^\mathcal{I}|(|\mathcal{V}^\mathcal{I}| + |\mathcal{E}^\mathcal{I}|)$ is due to checking the corresponding paths on every node in the network. However, considering the tripartite structure of ICI networks and since every iteration of the M-H algorithm only adds or removes one edge, Theorem \ref{theorem1} guarantees that constraints \ref{constraint1}--\ref{constraint6} are validated in \textit{linear} time.

\begin{theorem}\label{theorem1}
Let $\overrightarrow{\bm{A}}', \overleftarrow{\bm{A}}'$ denote the adjacency list and the reversed adjacency list of the proposed graph $\bm{G}'$ after adding or removing the edge $(k,j)$ selected from the feasible set $\mathcal{S}$. Given that the graph $\bm{G}$ in the last iteration of the M-H algorithm has already satisfied topological constraints \ref{constraint1}--\ref{constraint10} and the current proposed graph $\bm{G}'$ has already satisfied constraints \ref{constraint7}--\ref{constraint8},  $\bm{G}'$ satisfies constraints \ref{constraint1}--\ref{constraint6}, \ref{constraint9}--\ref{constraint10} if and only if $\overrightarrow{\bm{A}}'_k\ne\emptyset$ and $\overleftarrow{\bm{A}}'_j\ne\emptyset$.
\end{theorem}
The proof of Theorem \ref{theorem1} is presented in Appendix \ref{sec:theorem1}. Theorem \ref{theorem1} indicates that we can validate the proposed topology by only checking the out-degree of the heading node $k$ and the in-degree of the tailing node $j$, which only takes linear time ${\mathcal{O}(|{\mathcal{V}^{{\mathcal M}}}| + |{\mathcal{E}^{{\mathcal M}}}| + |{\mathcal{V}^{{\mathcal I}}}| + |{\mathcal{E}^{{\mathcal I}}}|)}$.

\subsection{Tie-No-Tie Sampler}\label{subsec: tnt}

Another technique to improve the sampling efficiency is to replace the \textit{random} sampler at line $10$ in Alg. $1$ with the \textit{`tie-no-tie'} (TNT) sampler \cite{gray2020bayesian}. The random sampler selects node pairs with equal probability, resulting in a higher frequency of proposing node pairs with non-edges (edges not present in the network) and adding the corresponding edges, then proposing node pairs with edges and removing the corresponding edges in sparse graphs. However, the newly added edges are likely to be rejected due to the violation of topological constraints \ref{constraint1}--\ref{constraint10} or the generation of a network topology discouraged by the cascading failure data. Conversely, the TNT sampler first selects the set of edges or the set of non-edges with equal probability, and then changes a random node pair in that set, which increases the frequency of removing edges in the proposal. Since the removed edges are more likely to be accepted due to graph sparsity, the sampling efficiency is improved.

\section{Numerical Experiments} \label{sec: results}

Here, we apply the proposed Bayesian approach to one synthetic ICI. We design five methods from different combinations of the IP and different computational optimization techniques (Table \ref{tab: method_comb}). By comparing the performance of these methods on reconstructing the synthetic system of ICI networks, we can identify the best method.

\begin{table}[H]
\caption{The configuration of five methods for ICIs reconstruction.}
\label{tab: method_comb}
\centering
\begin{tabular}{|c|c|c|c|c|}
\hline
\diagbox{\textrm{Method}}{\textrm{Technique}}&
\textrm{IP}&
\textrm{TNT}&
\textrm{Edge~List}&
\textrm{Validation}
\\
\hline
\multicolumn{5}{|c|}{With the IP} \\
\hline
M-1 & \checkmark & \checkmark & \checkmark & \checkmark\\
\hline
M-2 & \checkmark & \checkmark & \checkmark &\\
\hline
M-3 & \checkmark & \checkmark &  & \\
\hline
M-4 & \checkmark &  &  & \\
\hline
\multicolumn{5}{|c|}{Without the IP} \\
\hline
M-5 &  &  &  & \\
\hline
\end{tabular}
\end{table}

\subsection{Network Reconstruction of the Synthetic Networks}

\begin{figure}[!ht]
\centering
\begin{subfigure}[b]{0.45\textwidth}
    \centering
    \includegraphics[width=0.9\linewidth]{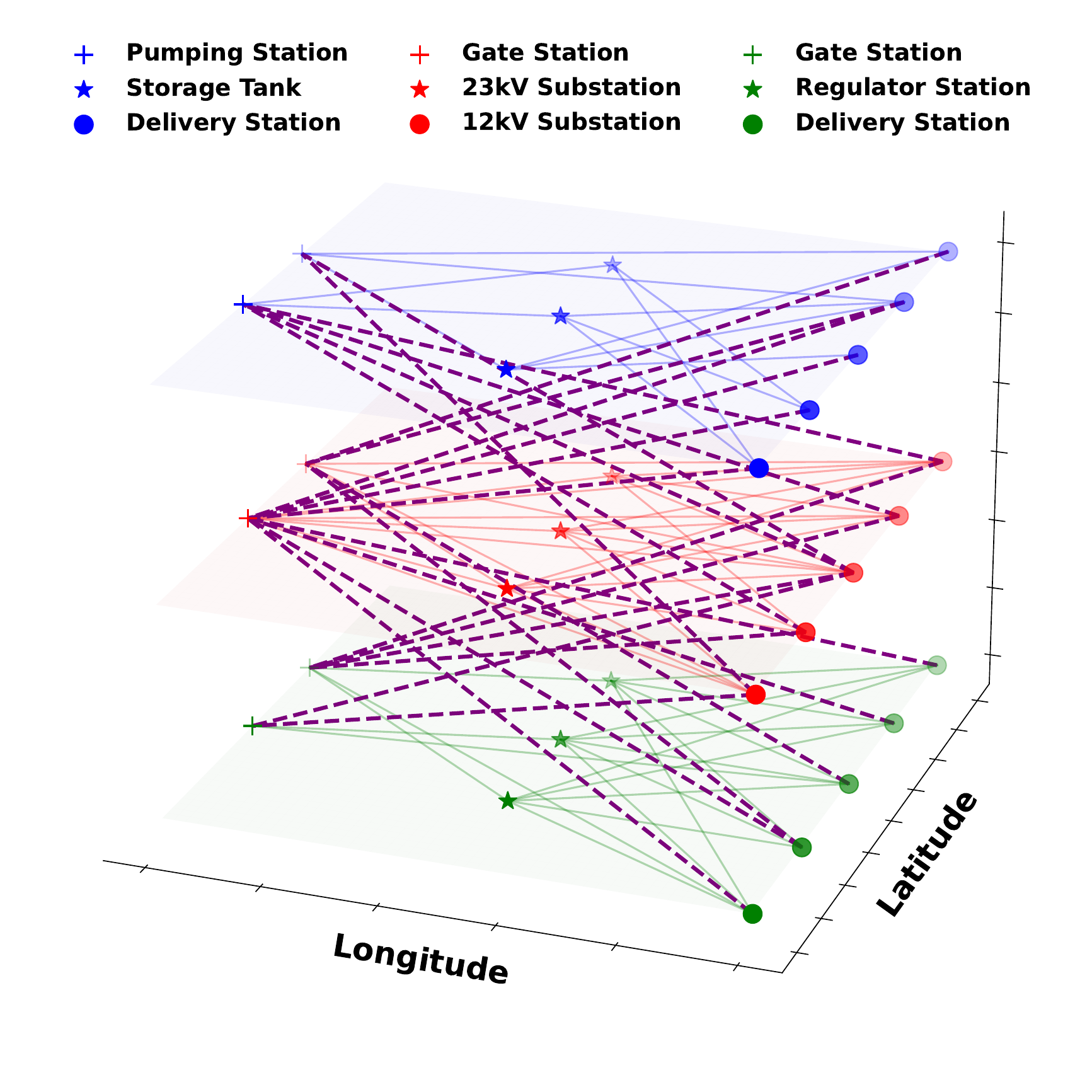}
    \caption{}
    \label{fig: case1network-a}
  \end{subfigure}
\begin{subfigure}[b]{0.45\textwidth}
\includegraphics[width=0.72\linewidth]{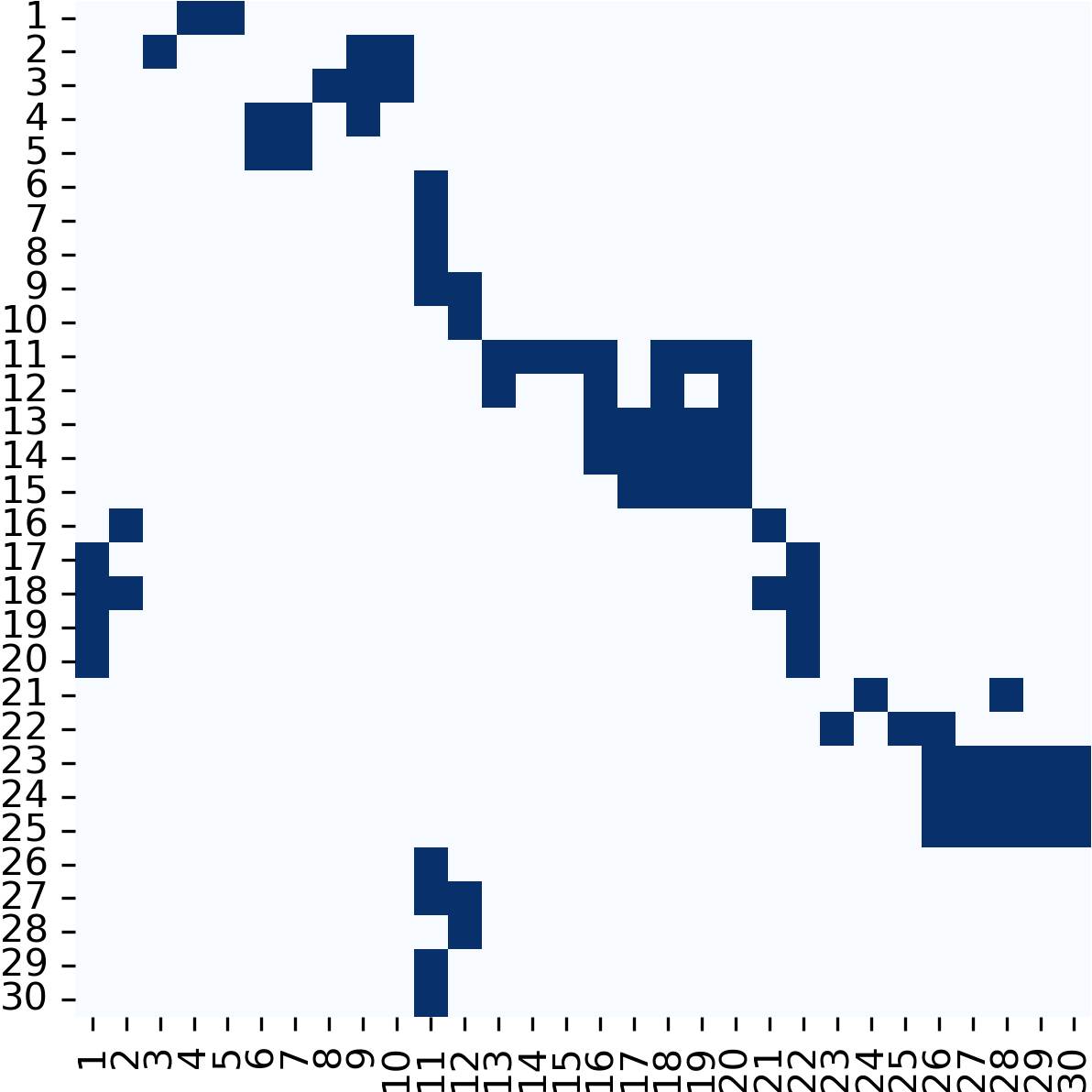}
\caption{}
\label{fig: case1network-b}
\end{subfigure}
\caption{(a) The water-power-gas (WPG) SICIN visualization. Solid lines represent edges within infrastructure networks, while dashed lines represent interdependent edges between infrastructure networks. Color code: \textcolor{blue}{Water}: blue, \textcolor{red}{Power}: red, and \textcolor{OliveGreen}{Gas}: green. (b) The adjacency matrix ($1$'s and $0$'s shown by dark and light blue squares, respectively) of the target synthetic network to be reconstructed for the WPG SICIN.}
\end{figure}

To assess the effectiveness of the proposed network reconstruction Bayesian approach, we primarily consider the class of synthetic networks. The synthetic network contains $3$ blocks corresponding to water (W), power (P), and gas (G) networks, each of which has $3$ hierarchical levels corresponding to supply \textit{s}, transmission \textit{t}, and demand \textit{d} facilities (Fig. \ref{fig: case1network-a}). All $3$ networks have $2$ supply nodes, $3$ transmission nodes, and $5$ demand nodes. The interdependency across networks is simulated using a previously developed approach for generating Synthetic ICI Networks (SICINs) \cite{wang2021generating}, which requires as input the degree distribution extracted from real-world infrastructures of the same type. This prior work \cite{wang2021generating} introduces a method for generating SICINs to overcome the research bottleneck caused by a lack of real-world infrastructure data. The SICIN framework utilizes simulation and nonlinear optimization. It employs a modified simulated annealing (MSA) algorithm to determine optimal asset placement within individual networks. Moreover, it uses a novel pseudo-tripartite graph algorithm to generate the complex links that connect different infrastructure systems. SICIN has been previously validated through a comparison with real-world infrastructure networks that led to the smallest difference in the adjacency matrices (DA). Particularly, for the baseline case of the Water-Power-Gas (WPG) ICI network in Shelby County, TN, USA, almost all properties (i.e., topological characteristics) of SICIN-based synthetic networks are the closest to the real ICI network of Shelby County, affirming the trustworthiness of the SICIN framework developed by \cite{wang2021generating}.

We consider $4$ types of interdependencies: $1$) pumping stations require electricity from $12$kV substations for pumping water from nearby rivers ${P \rightarrow W}$, $2$) power gate stations rely on water from water delivery stations for cooling purposes ${W \rightarrow P}$, $3$) gas gate stations require electricity from $12$kV substations for extracting natural gas from underground ${P \rightarrow G}$, and $4$) power gate stations depend on natural gas from gas delivery stations for generating electricity ${G \rightarrow P}$. We denote the water-power-gas (WPG) SICIN to be reconstructed as $\bm{A}^{\ast}$ hereafter. In summary, for the WPG SICIN, we have $\mathcal{M} = \{W,P,G\}$ and $\mathcal{I} = \{P \rightarrow W, W \rightarrow P, P \rightarrow G, G \rightarrow P\}$.

The cascading failure data $\bm{C}$ is simulated using the SI epidemic model, the ratio of the initial failed nodes $f$ is set to $0.2$, and the failure propagation probability $q$ is set to $0.4$ to get sufficient cascading failure data, aligned with prior work in the literature \cite{gray2020bayesian}. We investigate the performance of the proposed approach considering different amounts of cascading failure data $\bm{C}$. We design $3$ experiments with at least $5$ time-steps each: (\textit{i}) $E_5^5$: $5$ cascading failure scenarios, (\textit{ii}) $E_5^{15}$: $15$ cascading failure scenarios, and (\textit{iii}) $E_5^{40}$: $40$ cascading failure scenarios. All three experiments are performed with and without the IPs.

\begin{figure*}[!ht]
\centering

\begin{subfigure}[b]{0.3\textwidth}
    \centering
    \includegraphics[width=0.9\textwidth]{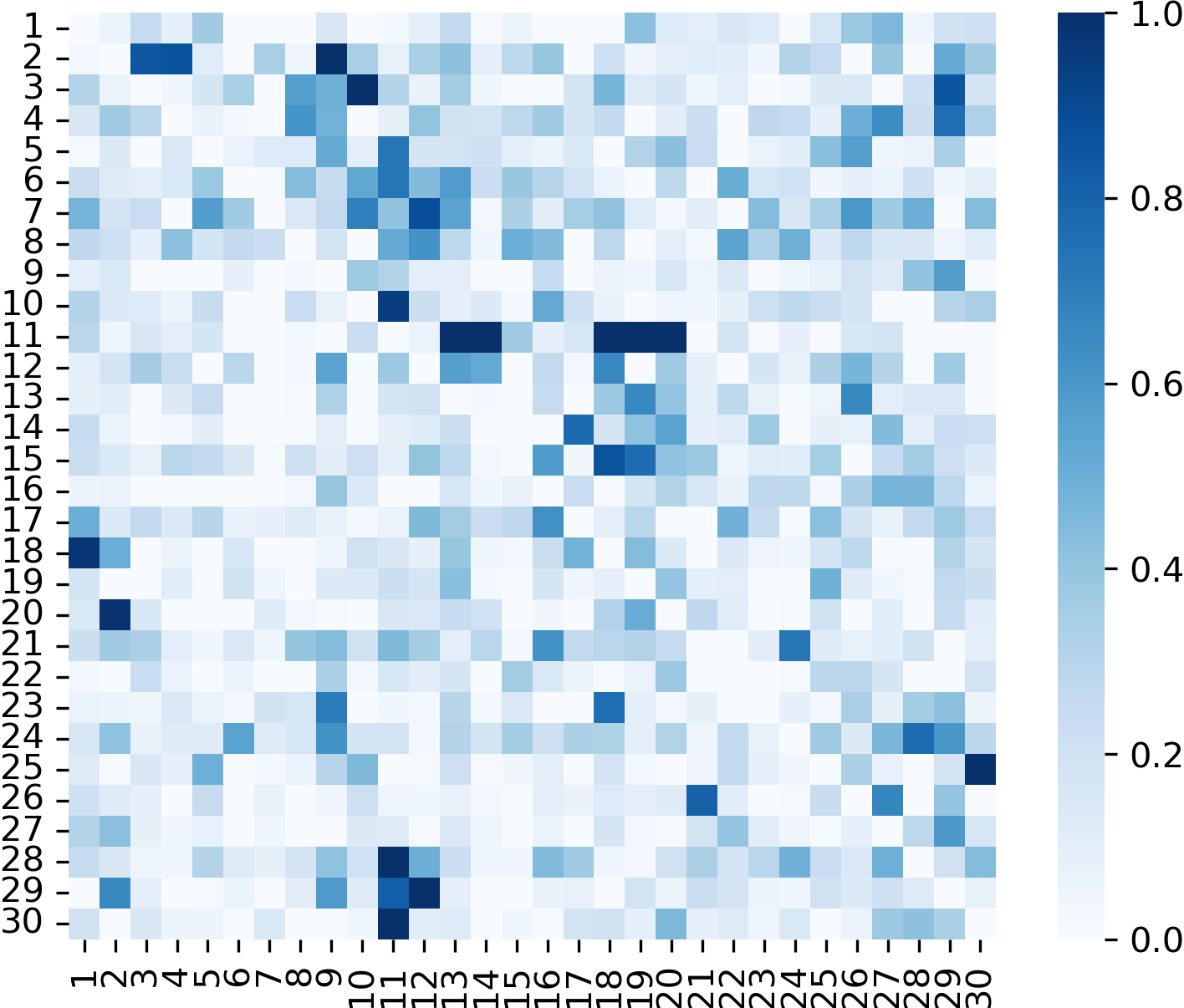}
    \caption{$E_5^5$, without the IP.}
    \label{fig1}
\end{subfigure}
\begin{subfigure}[b]{0.3\textwidth}
    \centering
    \includegraphics[width=0.9\textwidth]{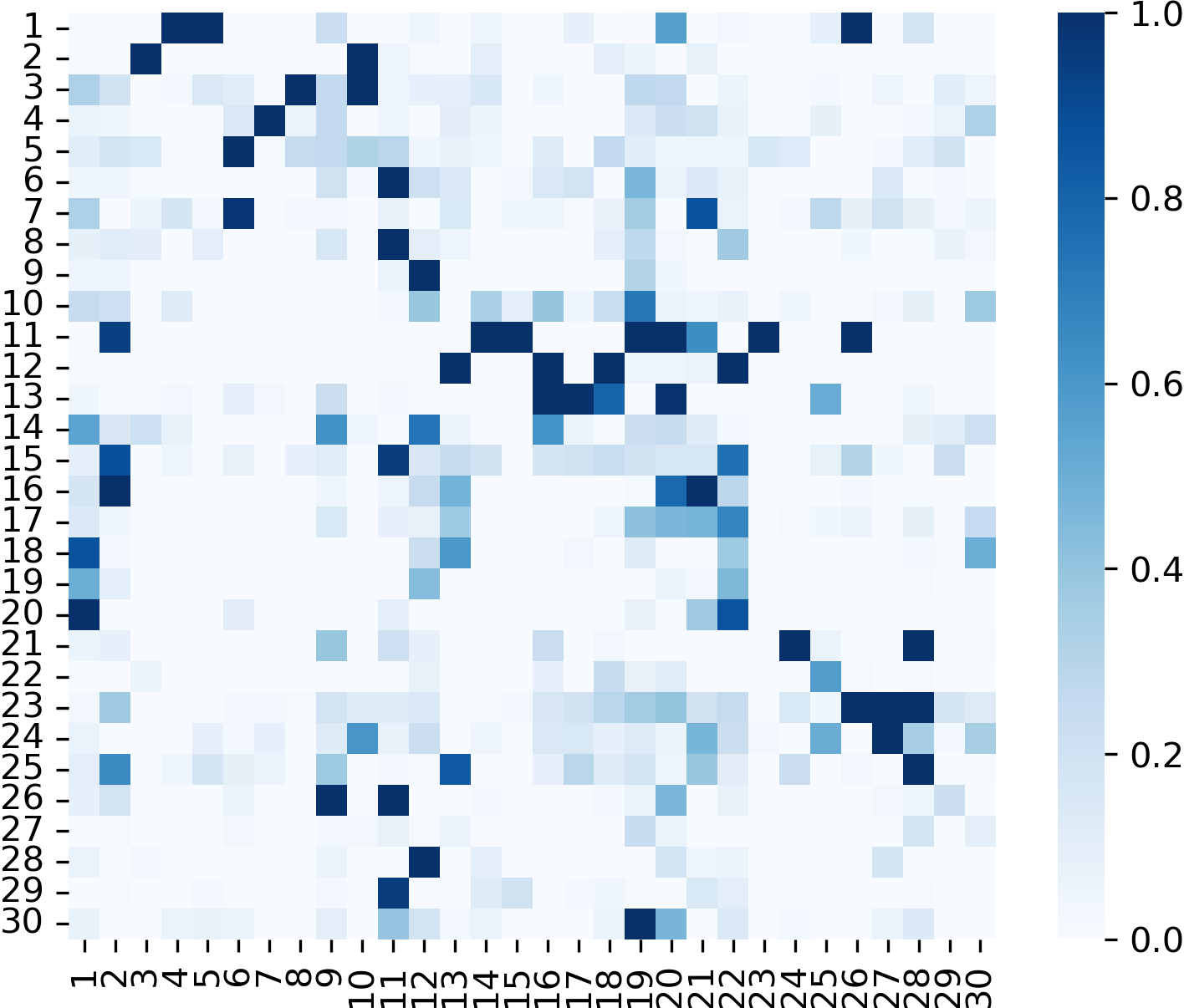}
    \caption{$E_5^{15}$, without the IP.}
    \label{fig2}
\end{subfigure}
\begin{subfigure}[b]{0.3\textwidth}
    \centering
    \includegraphics[width=0.9\textwidth]{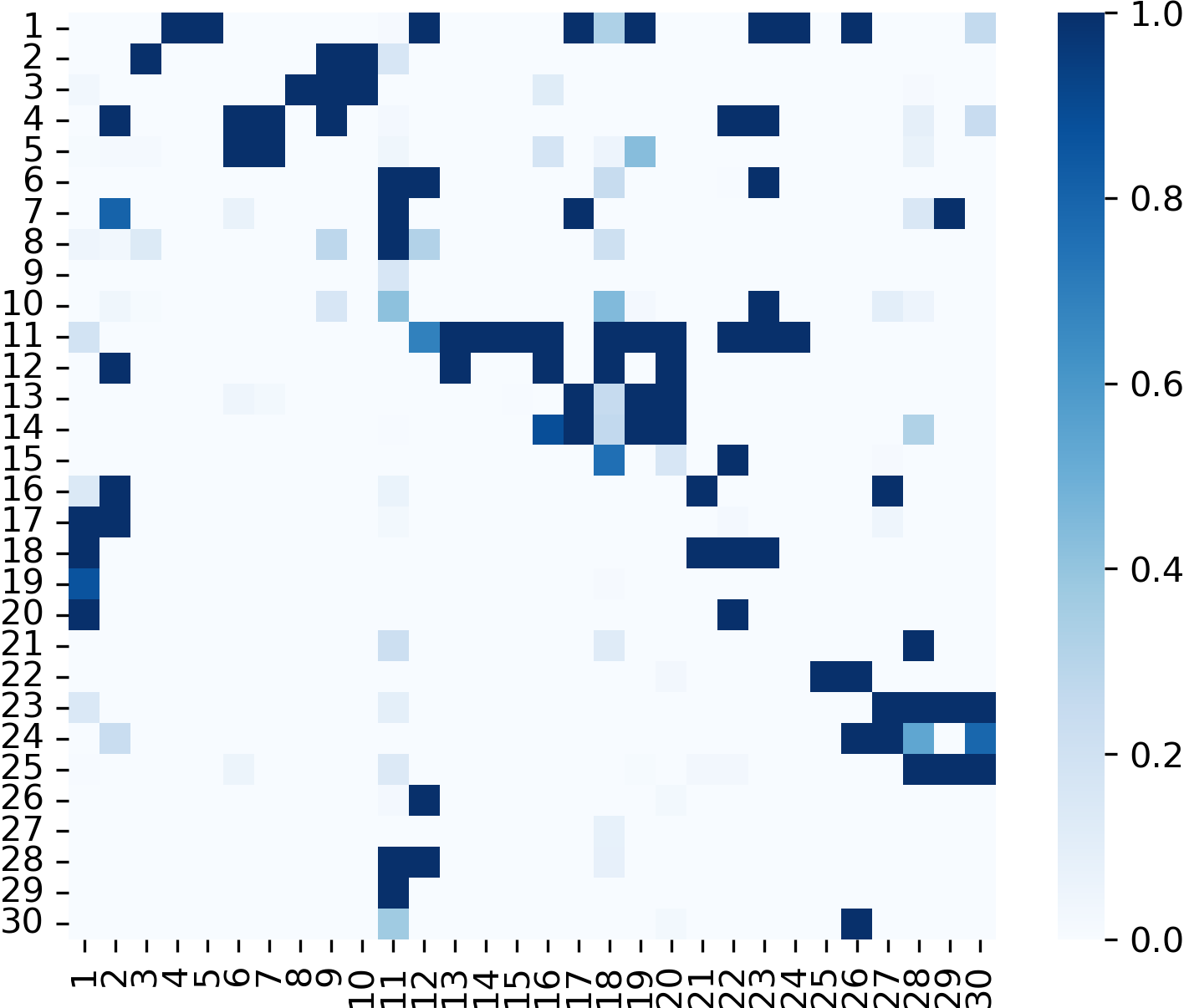}
    \caption{$E_5^{40}$, without the IP.}
    \label{fig3}
\end{subfigure}

\begin{subfigure}[b]{0.3\textwidth}
    \centering
    \includegraphics[width=0.9\textwidth]{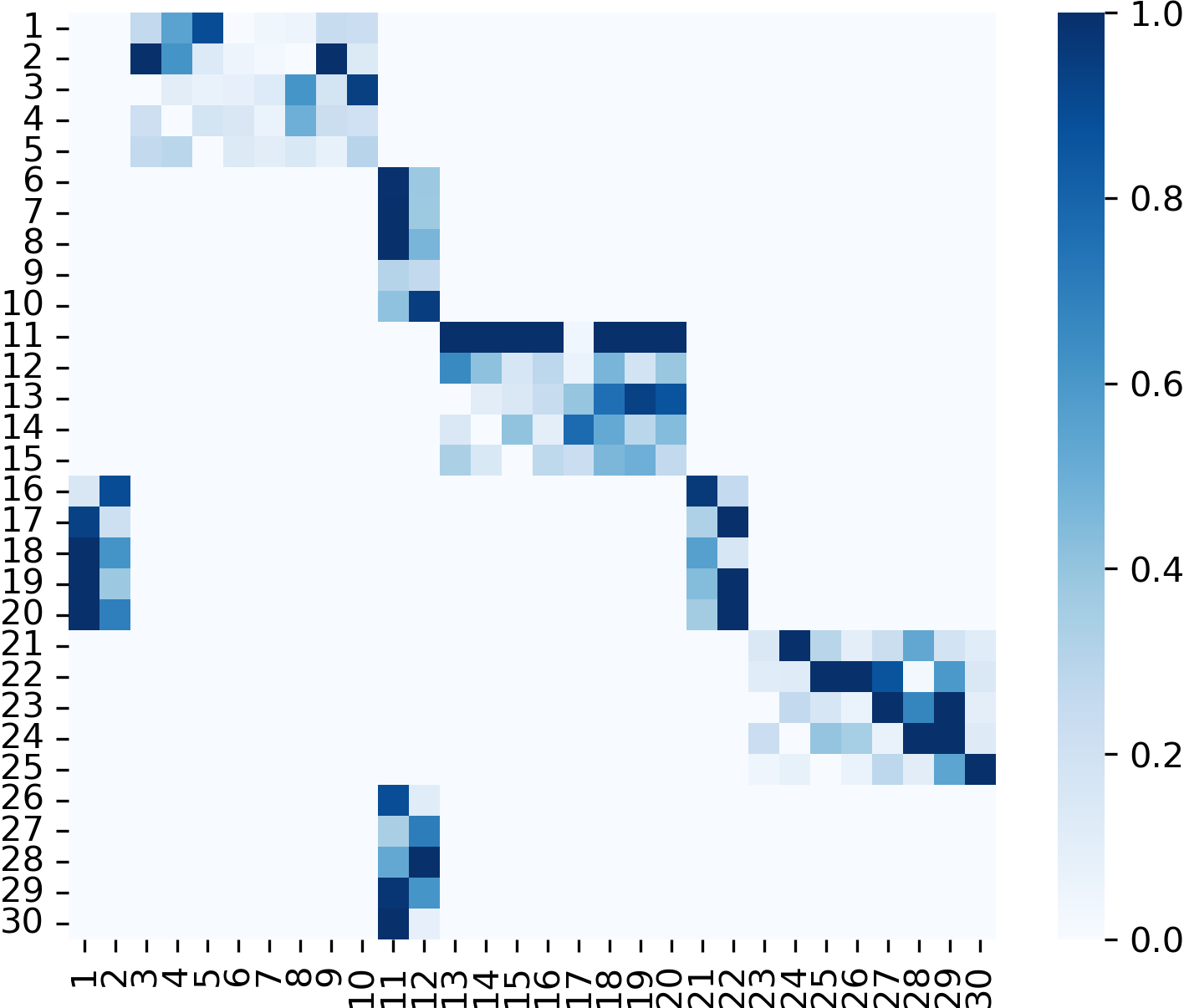}
    \caption{$E_5^5$, with the IP.}
    \label{fig4}
\end{subfigure}
\begin{subfigure}[b]{0.3\textwidth}
    \centering
    \includegraphics[width=0.9\textwidth]{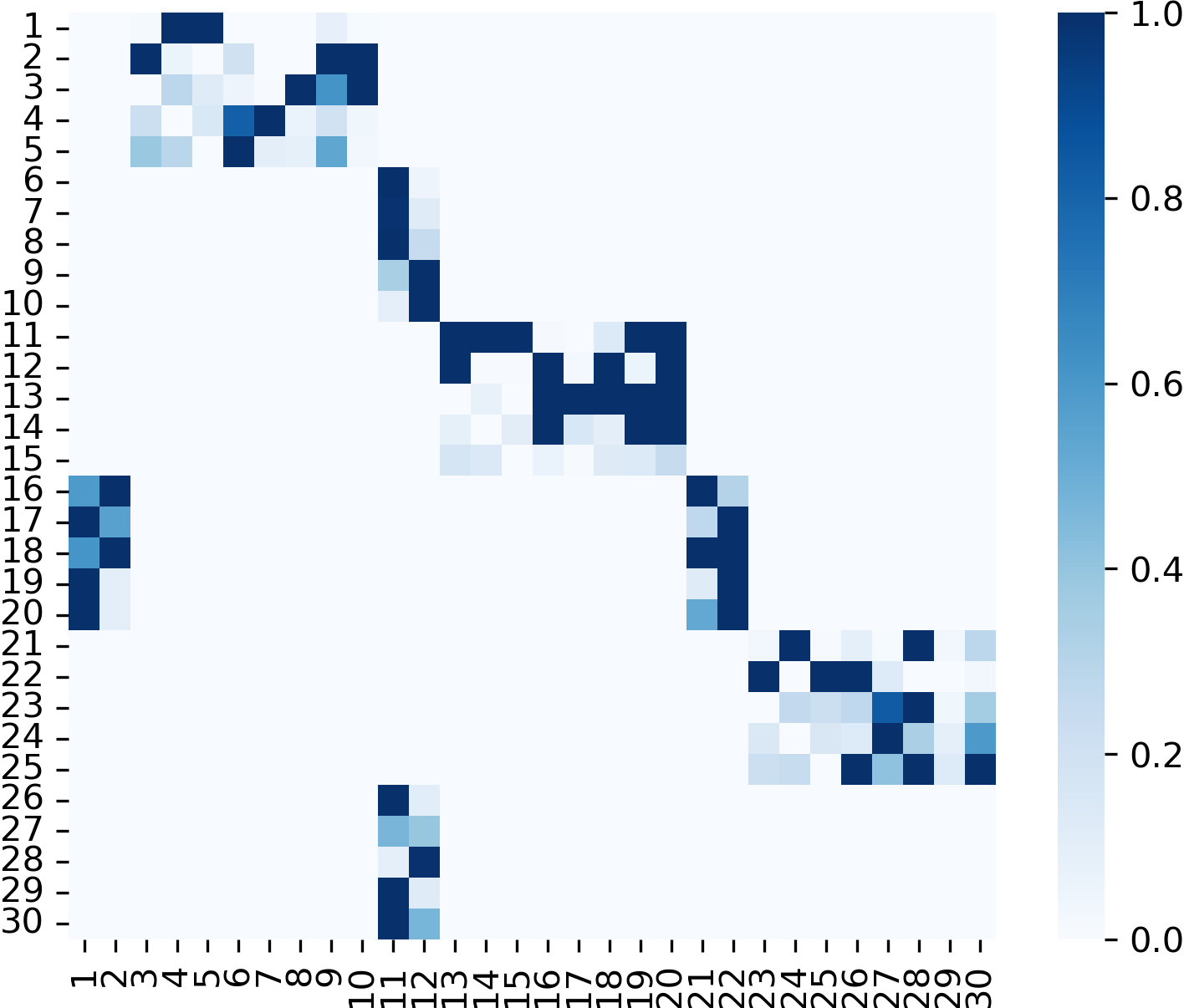}
    \caption{$E_5^{15}$, with the IP.}
    \label{fig5}
\end{subfigure}
\begin{subfigure}[b]{0.3\textwidth}
    \centering
    \includegraphics[width=0.9\textwidth]{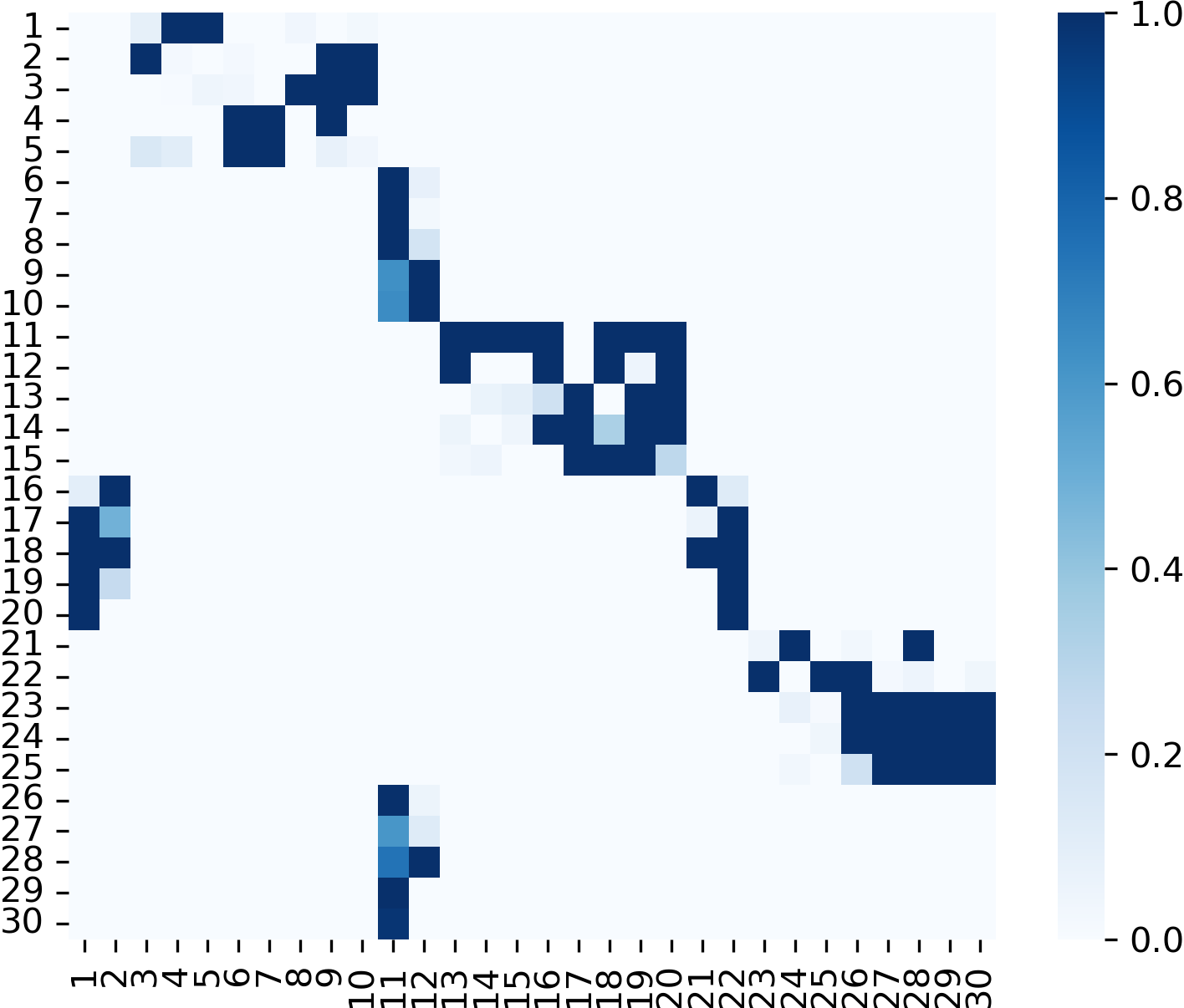}
    \caption{$E_5^{40}$, with the IP.}
    \label{fig6}
\end{subfigure}

\caption{The heatmap of the adjacency matrix reconstructed (a)--(c) without the IP (Configuration: TNT + Edge List), and (d)--(f) with the IP (Configuration: IP + TNT + Edge List). Three experiments for the WPG SICIN are considered as follows: (a), (d) $E_5^5$; (b), (e) $E_5^{15}$; (c), (f) $E_5^{40}$.}
\label{fig: heatmap}
\end{figure*}

We evaluate the accuracy of network reconstruction by considering it as a classification problem. Suppose that the set of networks in the discrete posterior distribution is $\mathcal{A}$, the edge probability $\hat{p}_{kj}$ for each node pair $(k,j)$ is defined as the ratio of the number of graphs with the edge between $k$ and $j$ in the posterior and the total number of graphs in the posterior as follows:
\begin{align} \label{eq:margin}
    & {\hat{p}_{kj}} = {|\{{\mathbf{A}\in \mathcal{A}}: \mathbf{A}_{kj} = 1\}|}/{|\mathcal{A}|}.
\end{align}The edge probability for each node pair in the network is displayed on the heatmap of the adjacency matrix in Fig. \ref{fig: heatmap}. In Figs. \ref{fig1}--\ref{fig3}, some false edges that are not present in the real networks are proposed (i.e., false positives) and shown in blank areas. In contrast, when using the IP (shown in Figs. \ref{fig4}--\ref{fig6}), the heatmap of the adjacency matrix conforms to the block structure in the original network (shown in Fig. \ref{fig: case1network-b}). Such a difference in reconstruction accuracy highlights the advantage of topological constraints \ref{constraint1}--\ref{constraint10} incorporation via the IPs. The heatmap of the adjacency matrix is darker in $E_5^{40}$ than in the other two experiments, indicating higher edge probability. Since $E_5^{40}$ uses more cascading failure data to update the network topology, we have higher confidence in the predicted edges.

\begin{figure}[!ht]
\centering
\begin{subfigure}[b]{0.45\textwidth}
    \centering
    \includegraphics[width=0.9\linewidth]{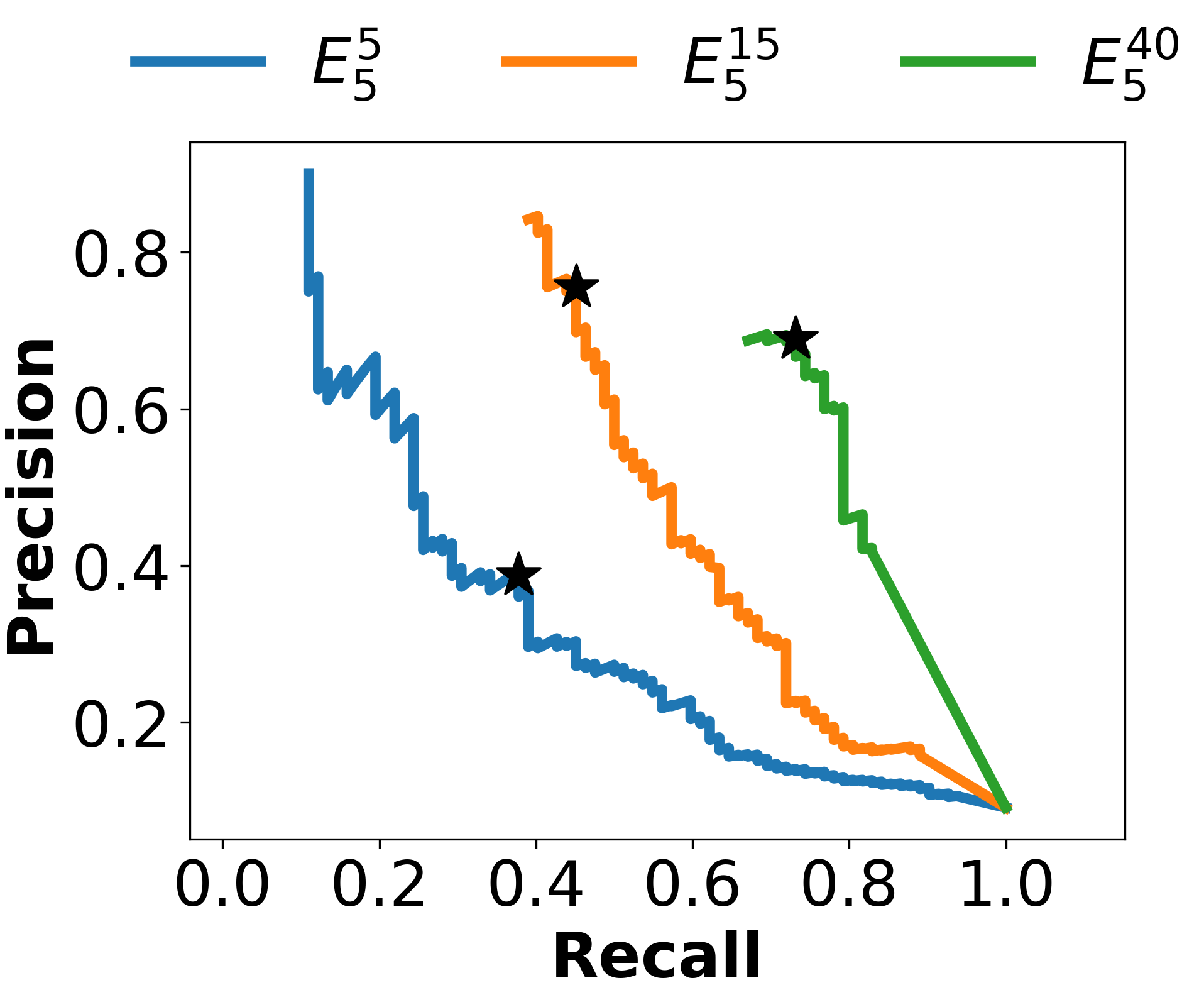}
    \caption{}
    \label{ff1}
\end{subfigure}
\begin{subfigure}[b]{0.45\textwidth}
    \centering
    \includegraphics[width=0.9\linewidth]{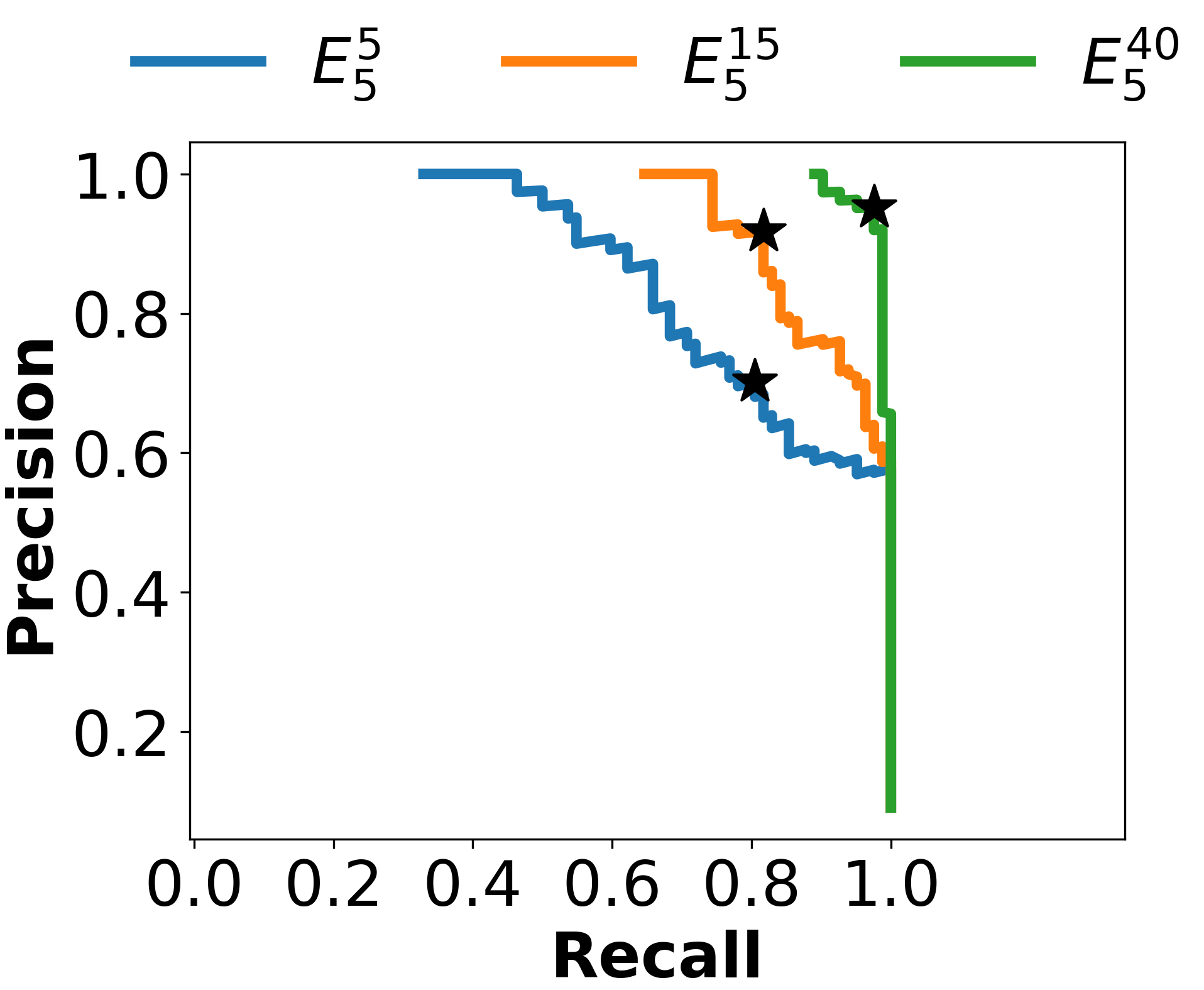}
    \caption{}
    \label{ff2}
\end{subfigure}
\caption{The precision-recall curves for the three experiments $E_5^{5}$, $E_5^{15}$, and $E_5^{40}$ (a) without the IP (Configuration: TNT + Edge List), and (b) with the IP (Configuration: IP + TNT + Edge List) for the WPG SICIN. The maximum F1-scores (pointed out with a $\star$ for each experiment) are: (a) $(0.383,0.565,0.710)$, and (b) $(0.750,0.865,0.964)$.}
\label{fig: precisionrecallcase1}
\end{figure}

\subsection{Performance of the Network Reconstruction Approach}

\subsubsection{Accuracy of the Network Reconstruction}

We now examine the accuracy of the proposed network reconstruction approach. We set a probability threshold $\rho$ and classify node pairs into two categories: node pairs connected by an edge if $\hat{p}_{kj} \ge \rho$ and node pairs that are not connected by an edge if $\hat{p}_{kj} < \rho$. We further count the number of edges for each node pair that are correctly classified
and calculate the \textit{F1-score}, which is the harmonic mean of \textit{precision} and \textit{recall}. The precision-recall curves are depicted in Fig. \ref{fig: precisionrecallcase1}. Precisely, for $10,001$ equidistant values of the probability threshold $\rho$ ranging from $0$ to $1$, we calculate the F1-scores and identify the maximum F1-score among those $10,001$ values. For the three experiments $E_5^{5}$, $E_5^{15}$, and $E_5^{40}$, the maximum F1-scores are $(0.383,0.565,0.710)$ for experiments without the IP (Configuration: TNT + Edge List), and $(0.750,0.865,0.964)$ for experiments with the IP (Configuration: IP + TNT + Edge List), respectively. The accuracy of the reconstructed network improves with additional cascading failure data. Also, the IP significantly improves the performance of reconstructing networks using the same amount of cascading failure data.

To further shed light on the network reconstruction accuracy of the IP-based method, we conduct comparisons with a graph-embedding-based network reconstruction baseline \cite{kipf2016semi,kipf2016variational}. Prior work proposes the auto-encoder framework, namely variational graph auto-encoder (VGAE), consisting of a graph convolutional network (GCN) encoder and a simple inner product decoder used to reconstruct the adjacency matrix \cite{kipf2016variational}. Precisely, using node features (e.g., node types or historical failure patterns), VGAE learns a latent representation for each node and predicts edges based on the similarity between these representations. Feeding the same cascading failure data $\bm{C}$ to the VGAE, we obtain the following maximum F1-scores for the three experiments $E_5^5$, $E_5^{15}$, and $E_5^{40}$: $(0.711,0.714,0.712)$. Comparing these values for the VGAE with those of the non-IP-based $(0.383,0.565,0.710)$ and the IP-based $(0.750,0.865,0.964)$ methods, we realize that the accuracy-based superiority is as follows: IP-based $>$ VGAE $>$ non-IP-based. 

\begin{figure*}[!ht]
\centering

\begin{subfigure}[b]{0.3\textwidth}
    \centering
    \includegraphics[width=0.9\textwidth]{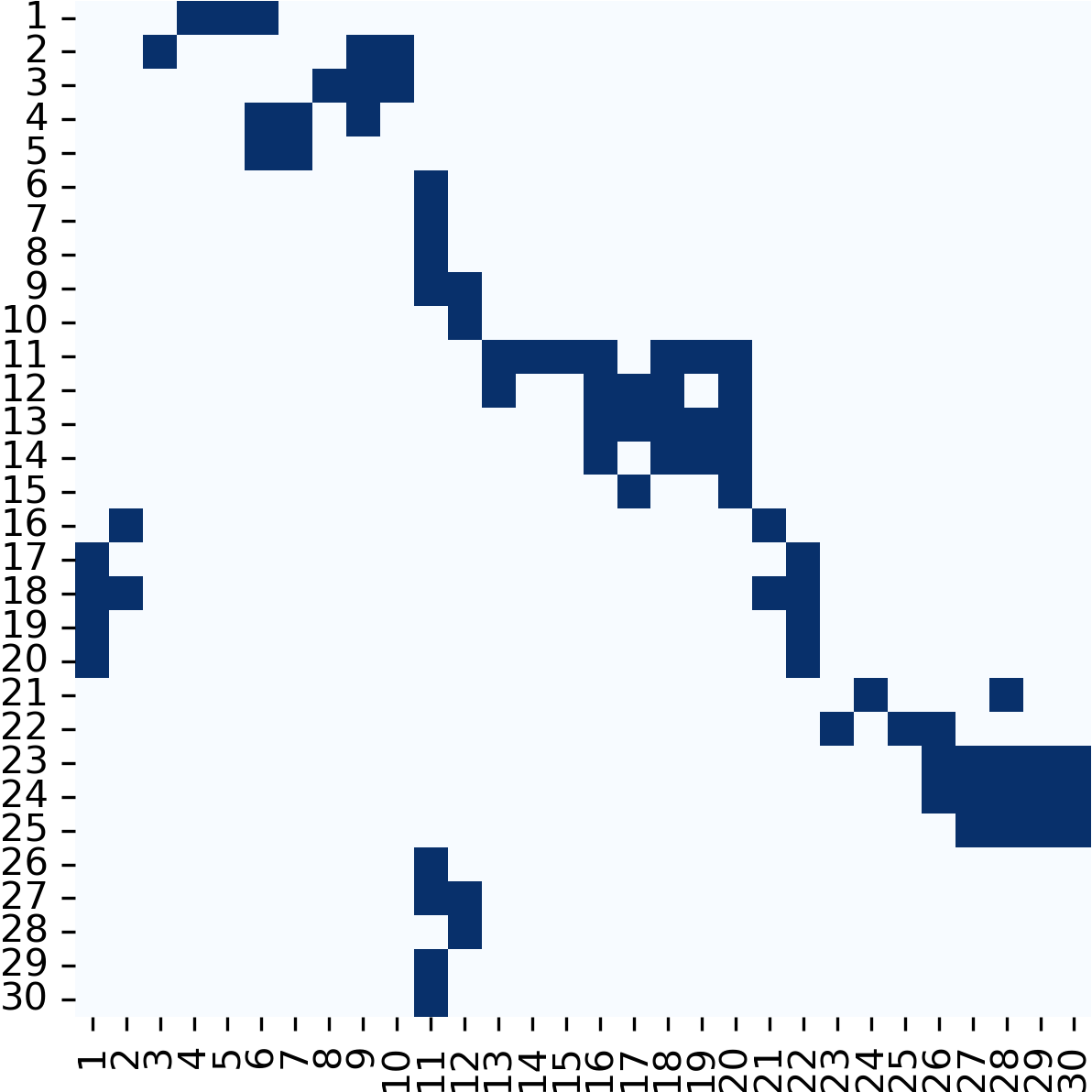}
    \caption{$E_5^{40}$, with the IP.}
    \label{fig01}
\end{subfigure}
\begin{subfigure}[b]{0.3\textwidth}
    \centering
    \includegraphics[width=0.9\textwidth]{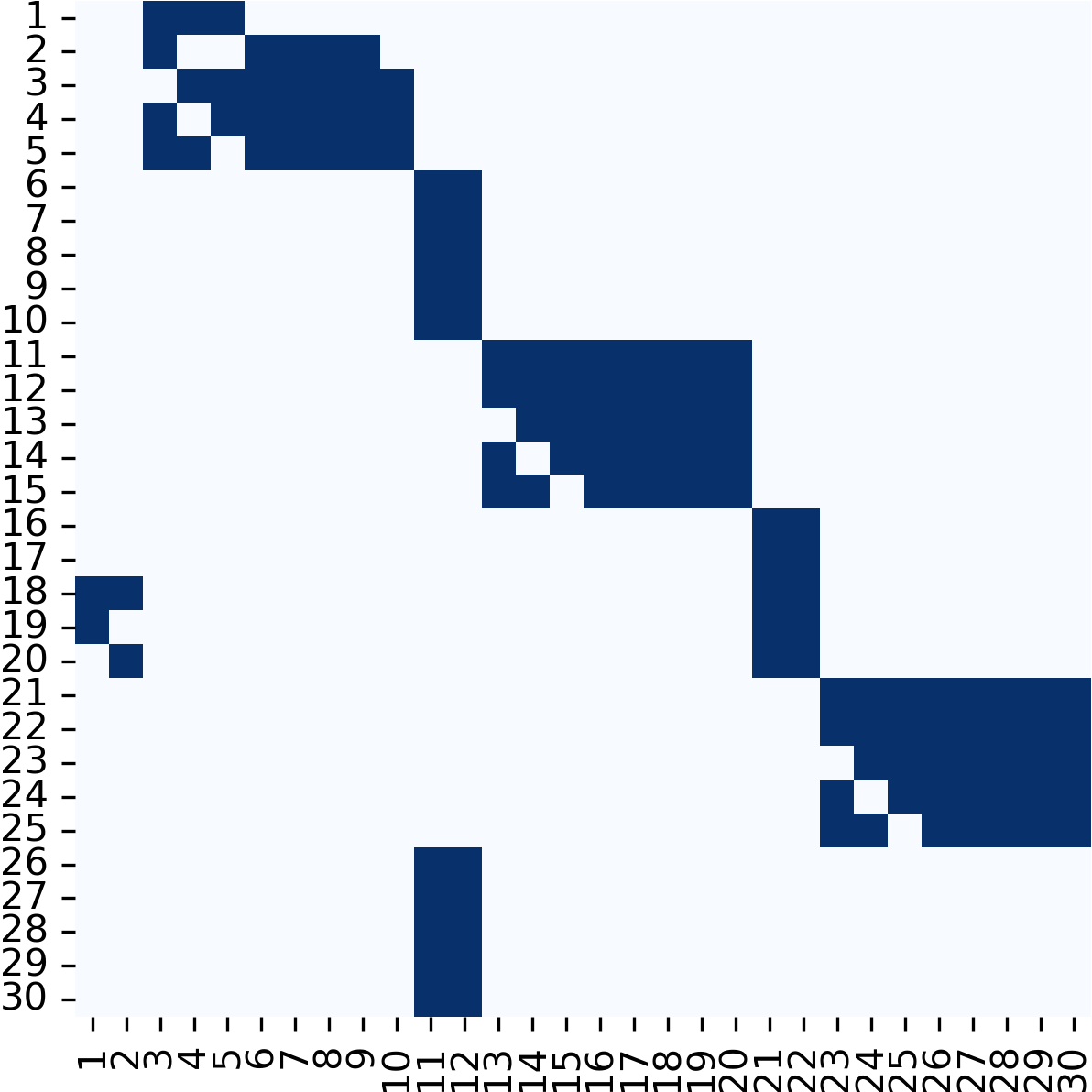}
    \caption{$E_5^{40}$, the VGAE.}
    \label{fig02}
\end{subfigure}
\begin{subfigure}[b]{0.3\textwidth}
    \centering
    \includegraphics[width=0.9\textwidth]{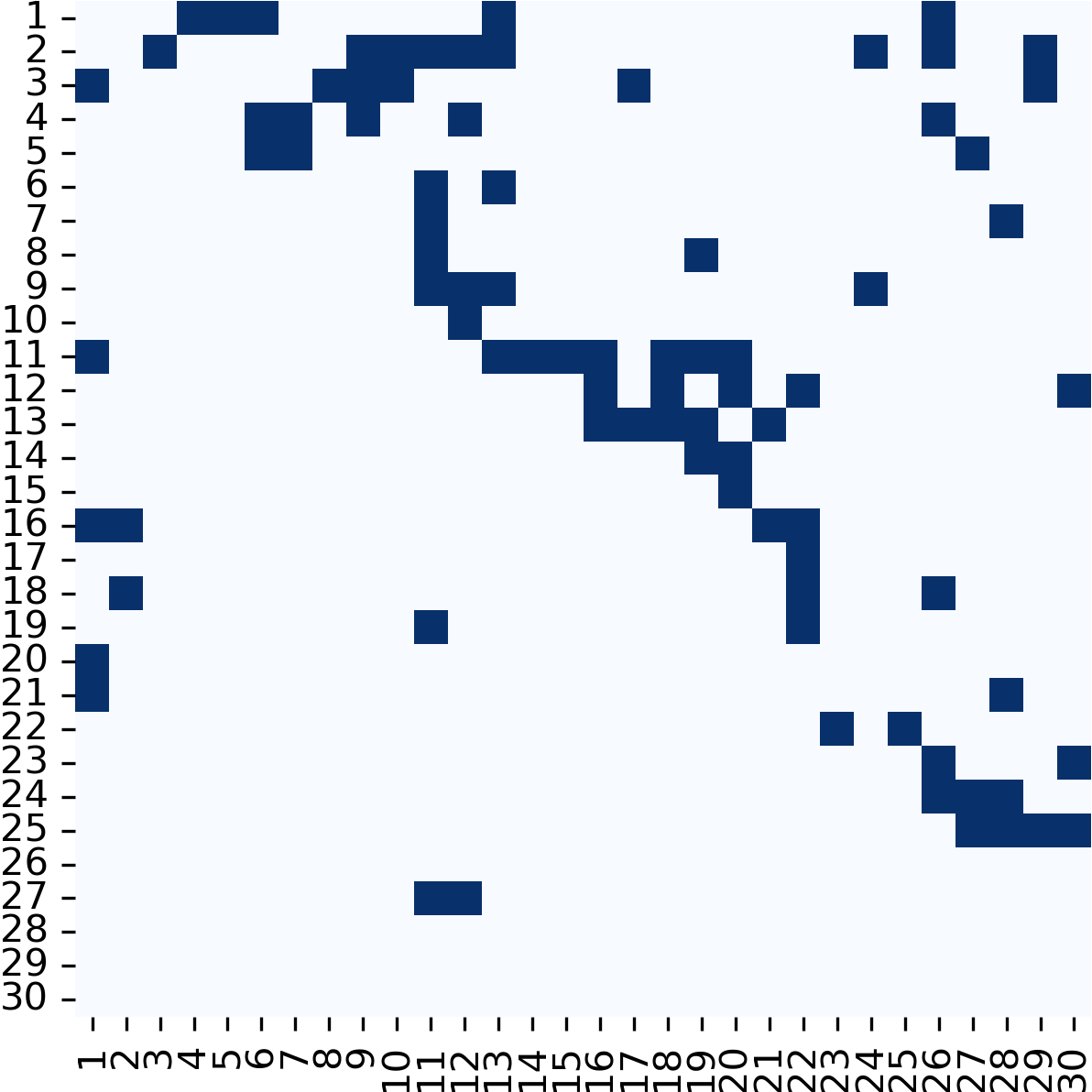}
    \caption{$E_5^{40}$, without the IP.}
    \label{fig03}
\end{subfigure}

\caption{The threshold $\rho$-based adjacency matrices reconstructed (a) with the IP (Configuration: IP + TNT + Edge List), (b) the VGAE, and (c) without the IP (Configuration: TNT + Edge List), considering the experiment $E_5^{40}$ for the WPG SICIN.}
\label{fig: comp}
\end{figure*}

Among the non-IP-based, the IP-based, and the VGAE, Fig. \ref{fig: comp} corroborates that the IP-based method is the only threshold $\rho$-based reconstructed network satisfying the topological constraints \ref{constraint1}--\ref{constraint10}. As mentioned earlier in Section \ref{sec: introduction}, the graph-embedding-based approaches like the VGAE \cite{kipf2016variational} require a massive amount of information to attain an acceptable reconstruction performance, while the Bayesian approaches like our IP-based method achieve an acceptable reconstruction performance given a relatively small amount of cascading failure data, which is driven by the specific nature of limited data availability for the ICI networks.

\begin{figure*}[!ht]
\centering

\begin{subfigure}[b]{0.3\textwidth}
    \centering
    \includegraphics[width=0.9\textwidth]{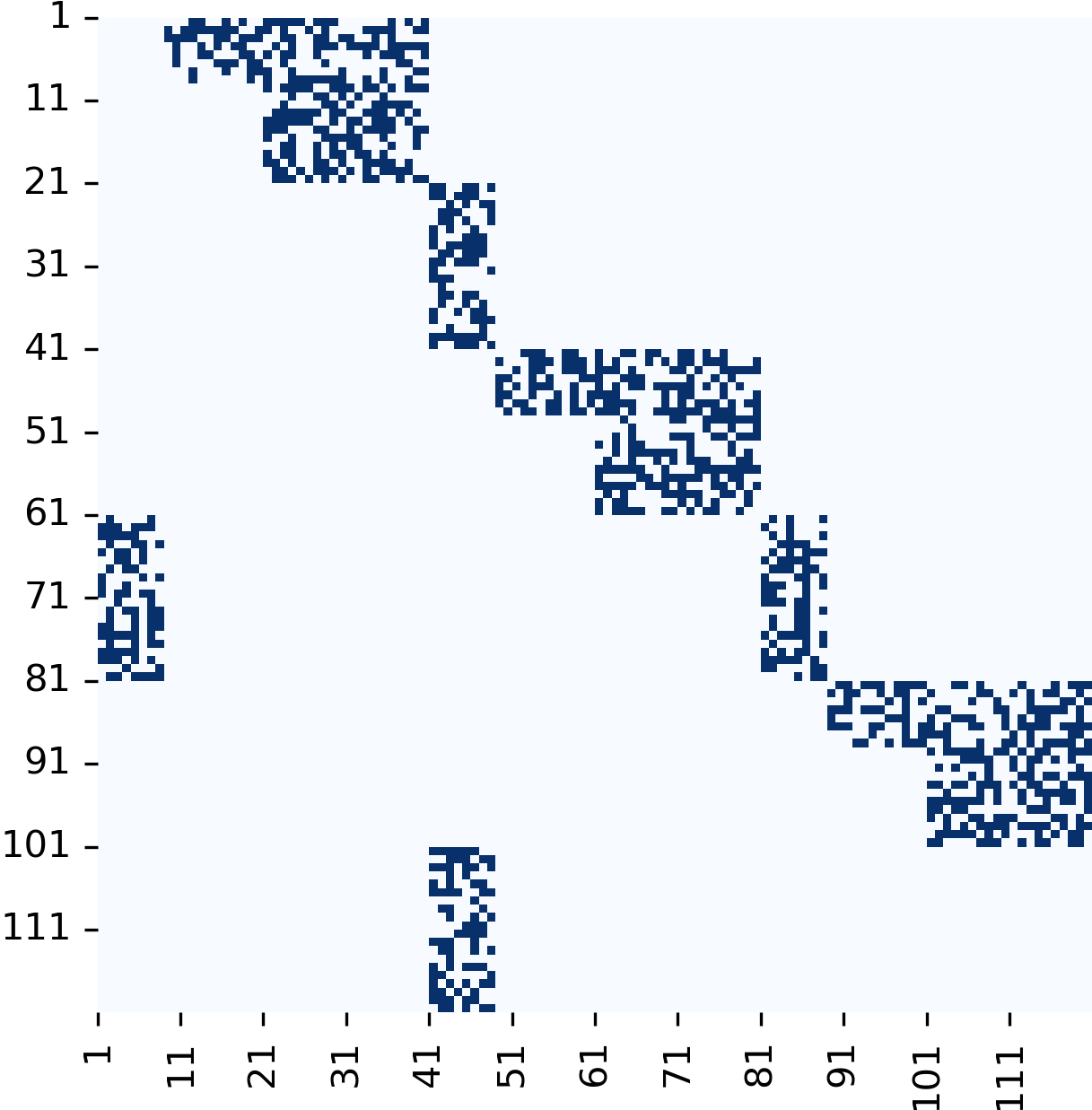}
    \caption{Target Network.}
    \label{ffig1}
\end{subfigure}
\begin{subfigure}[b]{0.353\textwidth}
    \centering
    \includegraphics[width=0.9\textwidth]{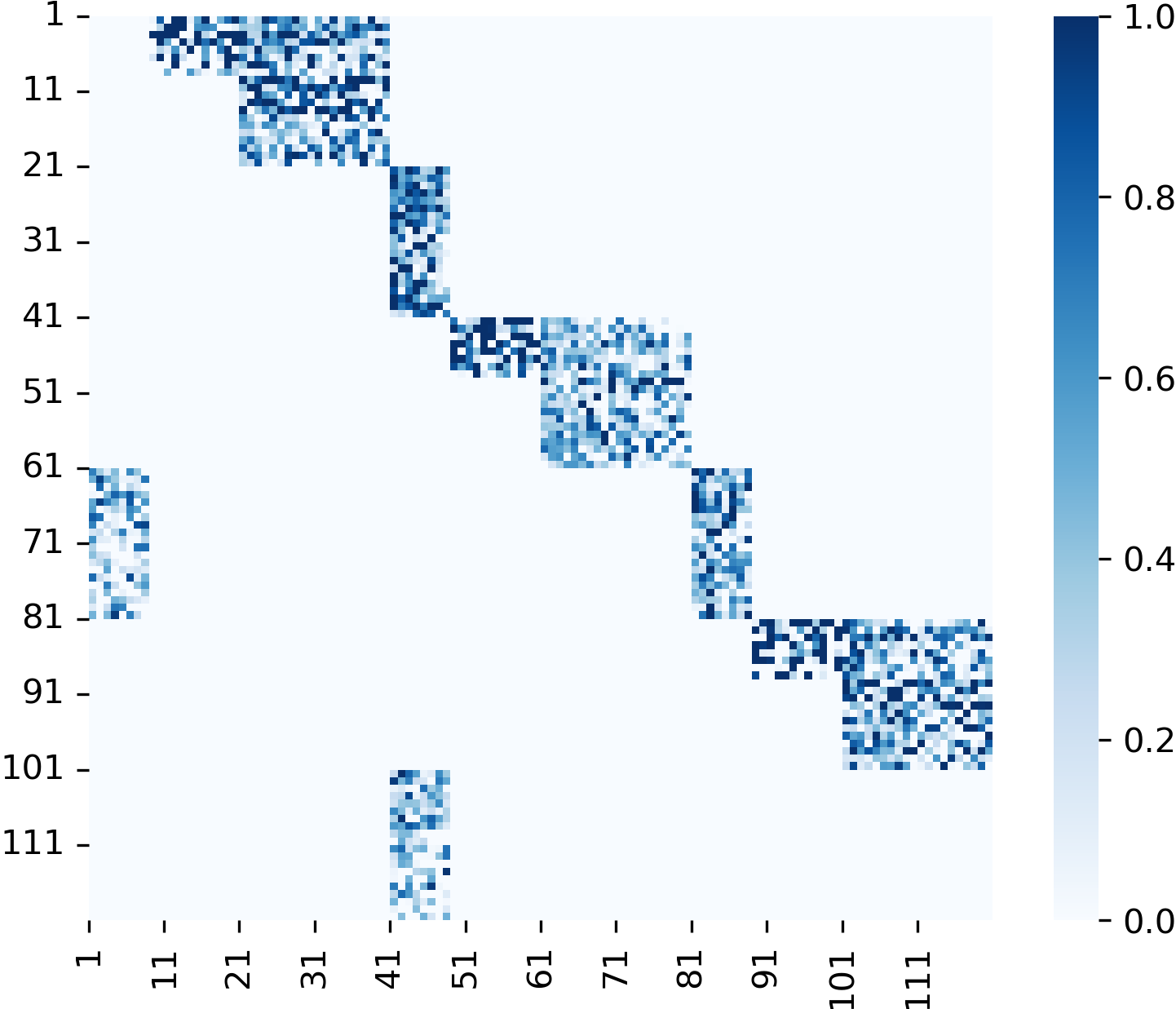}
    \caption{$E_4^{40}$, Reconstructed Network Heatmap.}
    \label{ffig2}
\end{subfigure}
\begin{subfigure}[b]{0.3\textwidth}
    \centering
    \includegraphics[width=0.9\textwidth]{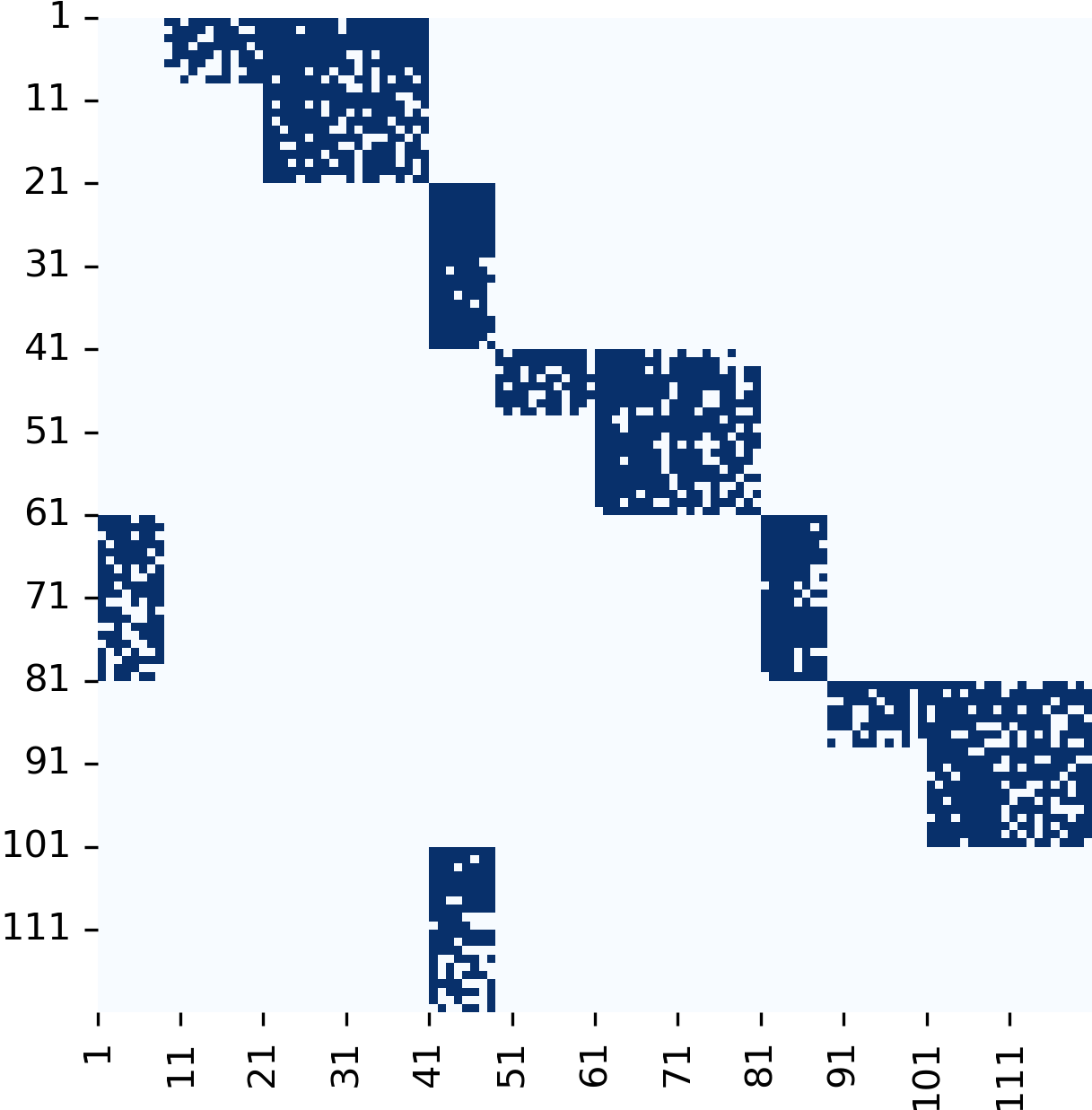}
    \caption{$E_4^{40}$, $\rho$-based Reconstruction.}
    \label{ffig3}
\end{subfigure}

\caption{(a) The adjacency matrix of the target network to be reconstructed for the $120 \times 120$ random WPG ICI network, (b) the heatmap of the reconstructed adjacency matrix with the IP (Configuration: IP + TNT + Edge List), and (c) the threshold $\rho$-based reconstructed adjacency matrix considering the experiment $E_4^{40}$.}
\label{ffig: heatmap}
\end{figure*}

\begin{figure}[!ht]
    \centering
    \includegraphics[width=0.9\linewidth]{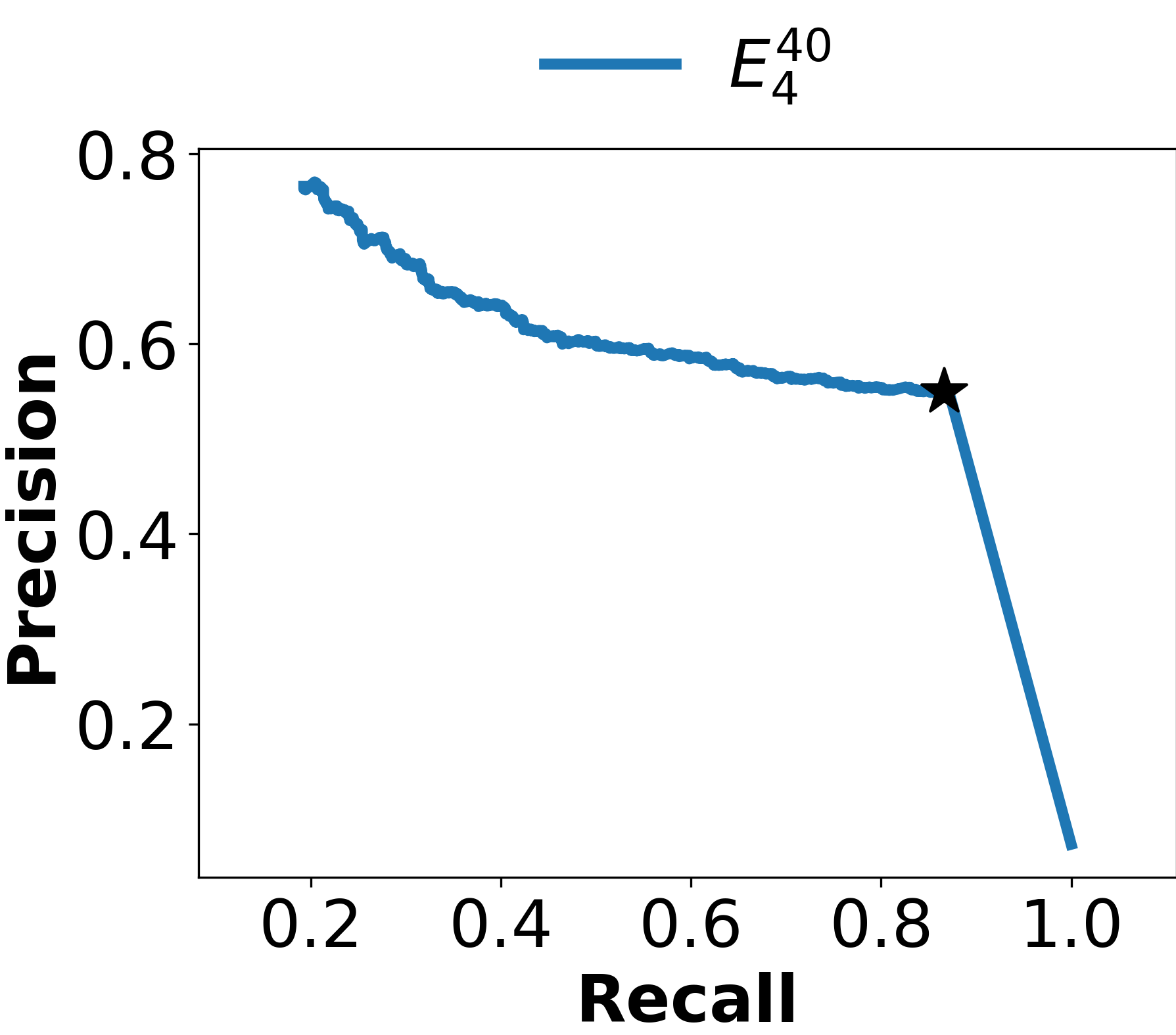}
    \caption{The precision-recall curve for the experiment $E_4^{40}$ with the IP (Configuration: IP + TNT + Edge List) for the $120 \times 120$ random WPG ICI network. The maximum F1-score (pointed out with a $\star$ for the experiment) and the running time (s) are $0.673$ and $217.025$, respectively.}
    \label{fffig:placeholder}
\end{figure}

To further verify the applicability of the IP-based method to larger ICI networks, we generate a $120 \times 120$ random WPG ICI network by setting edge probability $p$ equal to $0.5$ for feasible edges, ensuring the satisfaction of topological constraints \ref{constraint1}--\ref{constraint10}. All $3$ networks have $8$ supply nodes, $12$ transmission nodes, and $20$ demand nodes. Similarly, the cascading failure data $\bm{C}$ is simulated using the SI epidemic model, the ratio of the initial failed nodes $f$ is set to $0.2$, and the failure propagation probability $q$ is set to $0.4$ to get sufficient cascading failure data. It is worth noting that generating the cascading failure data $\bm{C}$ for larger ICI networks takes a longer time. Feeding the generated cascading failure data $\bm{C}$ to Alg. $1$ for an experiment with $40$ cascading failure scenarios and with $4$ time-steps, namely $E_{4}^{40}$, we obtain the network reconstruction results in Fig. \ref{ffig: heatmap}. Accordingly, Fig. \ref{fffig:placeholder} depicts the precision-recall curve associated with the network reconstruction in Fig. \ref{ffig: heatmap}.

Repeating the process for $10$ replicates, the average maximum F1-score and the average running time (s) are $0.664$ and $223.004$, respectively. The corresponding $95\%$ confidence intervals (CIs) are $[0.660, 0.668]$ and $[211.969, 234.040]$, respectively, indicating the high confidence for the calculated average values. For relatively dense (due to $p = 0.5$) and large-scale ICI networks, increasing the number of cascading failure scenarios and the failure length at the expense of higher computational cost could lead to an improvement in the achieved fair/moderate network reconstruction accuracy (i.e., F1-score = $0.664$).

\subsubsection{Effect of the Ratio of the Initial Failed Nodes}

\begin{figure}
    \centering
    \begin{subfigure}[b]{0.45\textwidth}
    \centering
    \includegraphics[width=0.9\linewidth]{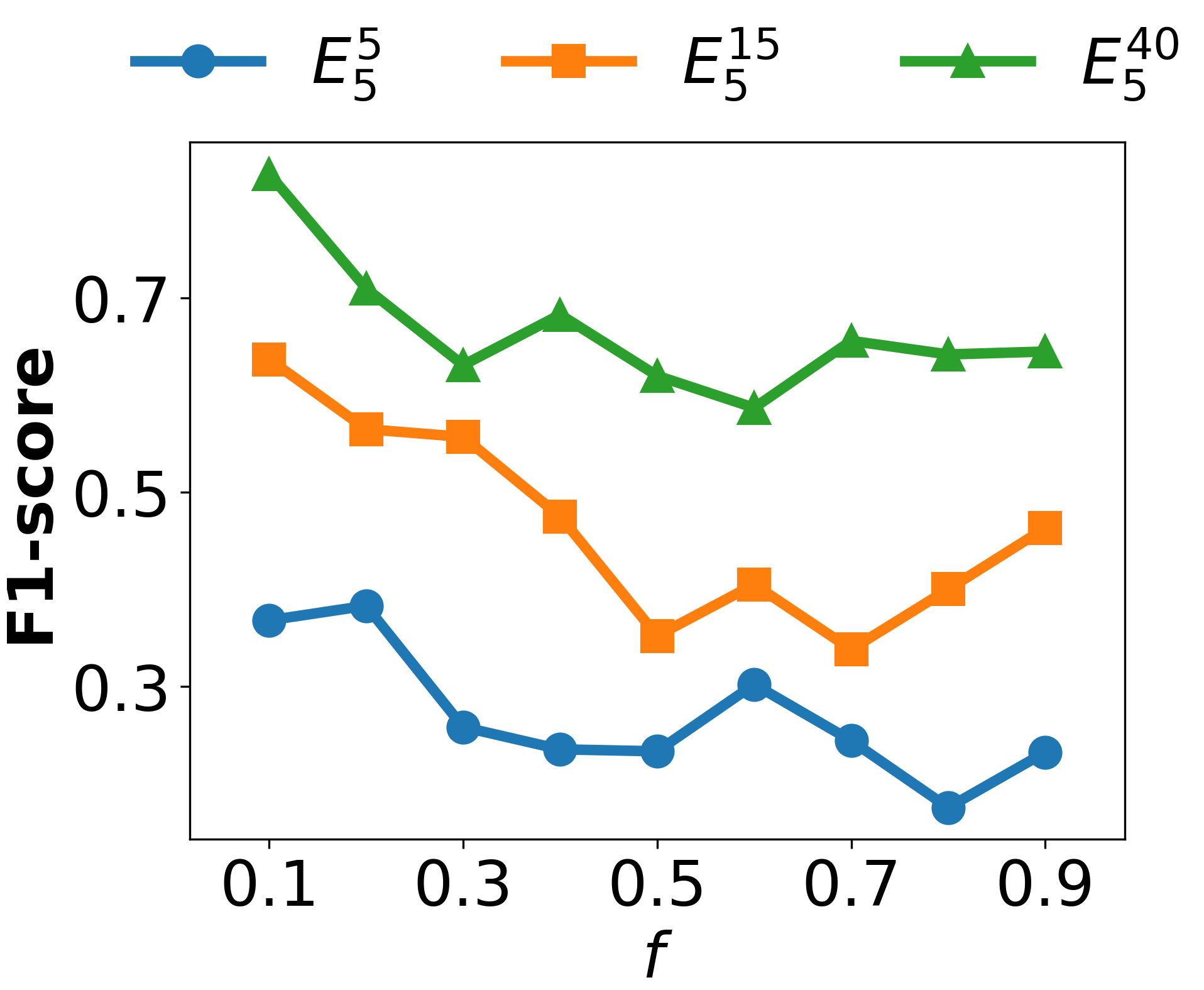}
    \caption{}
    \label{fg1}
    \end{subfigure}
    \begin{subfigure}[b]{0.45\textwidth}
    \centering
    \includegraphics[width=0.9\linewidth]{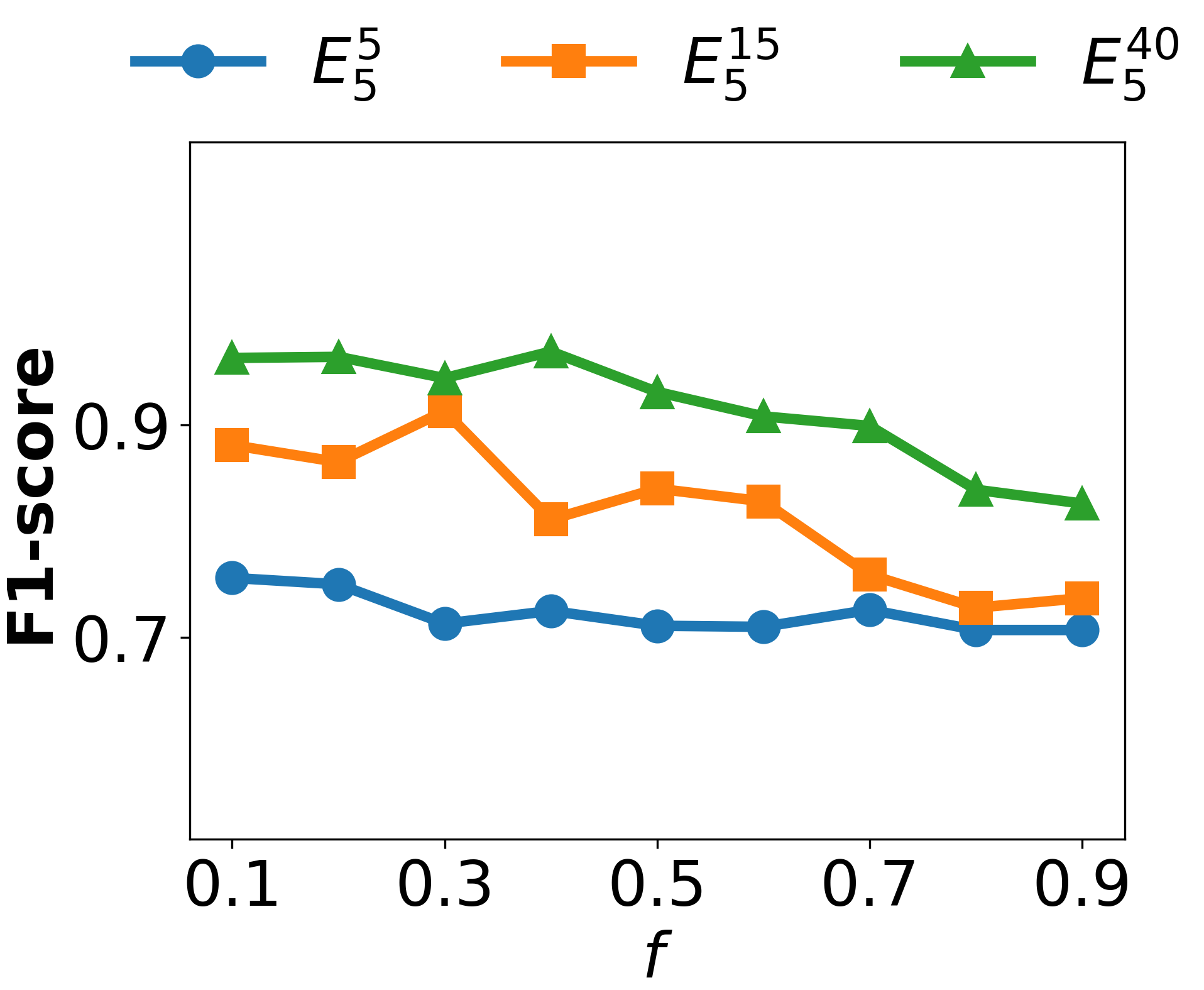}
    \caption{}
    \label{fg2}
    \end{subfigure}
    \caption{The relationship between the ratio of the initial failed nodes $f$ and the maximum F1-score for the three experiments $E_5^{5}$, $E_5^{15}$, and $E_5^{40}$ (a) without the IP (Configuration: TNT + Edge List), and (b) with the IP (Configuration: IP + TNT + Edge List) for the WPG SICIN.}
    \label{fig:placeholder}
\end{figure}

To investigate the effect of the ratio of the initial failed nodes $f$ on the network reconstruction accuracy (quantified by the maximum F1-score), we repeat the simulations with distinctly generated cascading failure data $\bm{C}$ by varying $f$ from $0.1$ to $0.9$ with a step size of $0.1$. The observations have been reflected in Fig. \ref{fig:placeholder} for the three experiments $E_5^{5}$, $E_5^{15}$, and $E_5^{40}$ without the IP (Configuration: TNT + Edge List) and with the IP (Configuration: IP + TNT + Edge List) for the WPG SICIN. As Fig. \ref{fig:placeholder} depicts, the moderate-value choices of $f \in \{0.2,0.3,0.4\}$ seem reasonable as such choices lead to better maximum F1-scores. It is particularly consistent with the choice of $f = 0.2$ in prior work in the literature \cite{gray2020bayesian}. The observations in Fig. \ref{fig:placeholder} further corroborate the fact that moderate values of $f$ ensure the high inferential precision by triggering a controlled, multi-branch cascade that propagates organically through the network's true paths.

On one hand, the low values of $f$ usually lead to high ambiguity due to the lack of data to differentiate between many candidate network topologies and failing to reach deep, isolated, or sparsely connected sections of the network. On the other hand, the high values of $f$ usually lead to significant loss of information due to overloading and suffering an immediate or near-instantaneous global collapse, weakening the effectiveness of the Bayesian inference. Moreover, comparing Fig. \ref{fg2} with Fig. \ref{fg1}, we realize that the magnitudes of relative sensitivity (i.e., elasticity) values of the reconstructions with the IP, $(0.036,0.130,0.098)$, are relatively less than those of the counterparts without the IP, $(0.349,0.304,0.123)$, respectively. In other words, the IP-based network reconstruction approach outperforms the non-IP-based counterpart by achieving a less sensitive network reconstruction (in terms of accuracy) against the ratio of the initial failed nodes.

\subsubsection{Effect of the Noise Level in the Cascading Failure Data}

\begin{figure}
    \centering
    \begin{subfigure}[b]{0.45\textwidth}
    \centering
    \includegraphics[width=0.9\linewidth]{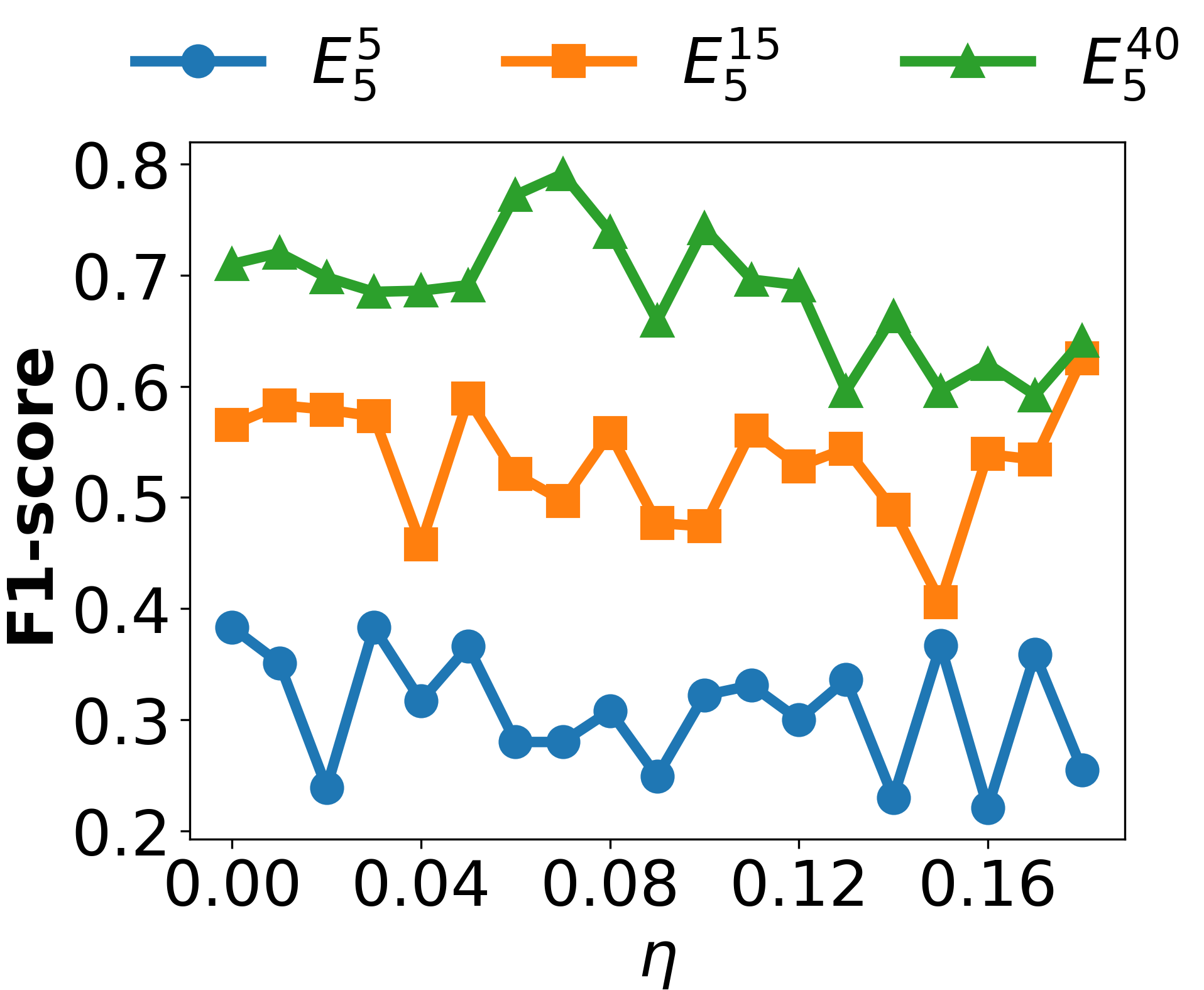}
    \caption{}
    \label{efg1}
    \end{subfigure}
    \begin{subfigure}[b]{0.45\textwidth}
    \centering
    \includegraphics[width=0.9\linewidth]{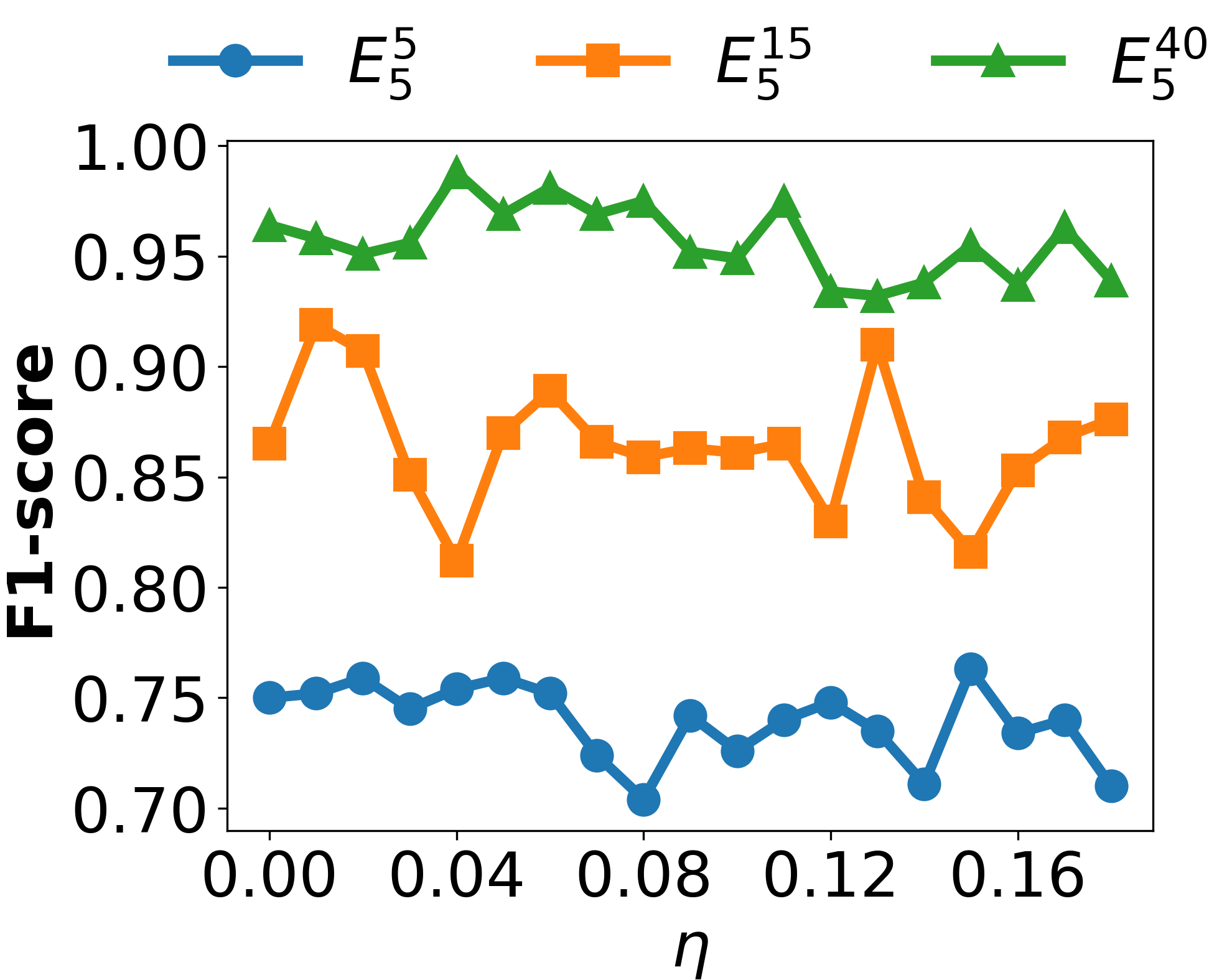}
    \caption{}
    \label{efg2}
    \end{subfigure}
    \caption{The relationship between the noise level $\eta$ and the maximum F1-score for the three experiments $E_5^{5}$, $E_5^{15}$, and $E_5^{40}$ (a) without the IP (Configuration: TNT + Edge List), and (b) with the IP (Configuration: IP + TNT + Edge List) for the WPG SICIN.}
    \label{efig:placeholder}
\end{figure}

To explore the effect of the noise level $\eta$ in the cascading failure data $\bm{C}$ on the network reconstruction accuracy (quantified by the maximum F1-score), for a fixed value of $f = 0.2$, we repeat the simulations with distinctly generated noisy cascading failure data $\bm{C}$ with different equidistant noise levels $\eta$'s ranging from $0$ (i.e., the noiseless baseline) to $0.18$ with a step size of $0.01$. By trial and error (e.g., a bisection-based search), we find that $0.18$ is the maximum tolerable noise level $\eta$ for which the network can be reconstructed. We have incorporated two classes of noises: \textit{(i)} missing data (i.e., dropouts) simulating a sensor failing to record a node's collapse, and \textit{(ii)} timing jitter (i.e., permutations) simulating logging delays where node $k$ is recorded as failing after node $j$, even if $k$ actually failed first. Through the noise level $\eta$, we will parametrically simulate observational noise (missing nodes or reporting delays) by altering the failure tracks immediately after they are generated but before they are used to initialize the prior. The observations have been visualized in Fig. \ref{efig:placeholder} for the three experiments $E_5^{5}$, $E_5^{15}$, and $E_5^{40}$ without the IP (Configuration: TNT + Edge List) and with the IP (Configuration: IP + TNT + Edge List) for the WPG SICIN. 

As Fig. \ref{efig:placeholder} demonstrates, an overall descending trend (i.e., trade-off) exists between the noise level $\eta$ and the maximum F1-score. Quantitatively, the Kendall tau values for the three experiments $E_5^{5}$, $E_5^{15}$, and $E_5^{40}$ are: $(-0.194,-0.193,-0.453)$ for the case without the IP in Fig. \ref{efg1} and $(-0.342,-0.152,-0.324)$ for the case with the IP in Fig. \ref{efg2}, respectively. A negative value of Kendall tau indicates the existence of an overall descending trend. Such a trade-off between the noise level $\eta$ and the maximum F1-score is intuitively aligned with our expectation regarding an adverse effect of the noise on the network reconstruction accuracy. Furthermore, comparing Fig. \ref{efg2} with Fig. \ref{efg1}, we realize that the magnitudes of relative sensitivity (i.e., elasticity) values of the reconstructions with the IP, $(0.017,0.014,0.014)$, are less than those of the counterparts without the IP, $(0.086,0.033,0.084)$, respectively. In other words, the IP-based network reconstruction approach outperforms the non-IP-based counterpart by achieving a less sensitive network reconstruction (in terms of accuracy) against the noise level.

\subsection{Computational Performance of the Three Incorporated Optimization Techniques}

\begin{figure}[!ht]
\centering

\begin{subfigure}[b]{0.45\textwidth}
    \centering
    \includegraphics[width=0.9\linewidth]{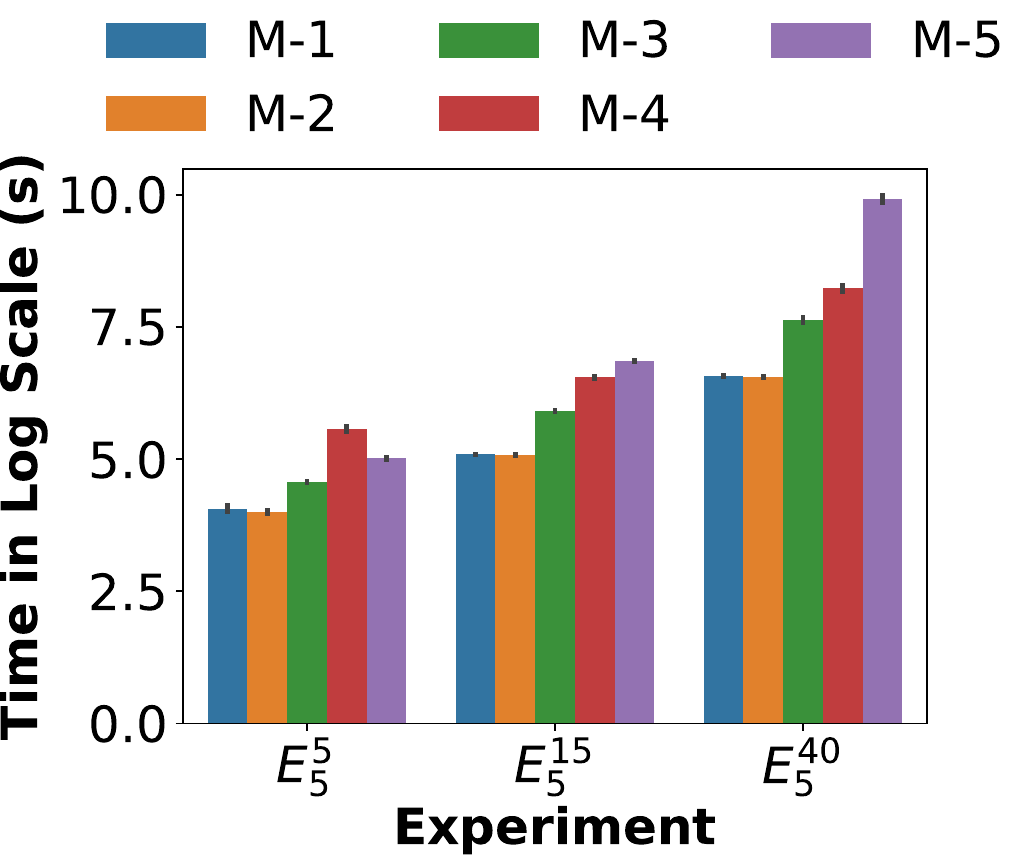}
    \caption{}
    \label{gg1}
\end{subfigure}
\begin{subfigure}[b]{0.45\textwidth}
    \centering
    \includegraphics[width=0.9\linewidth]{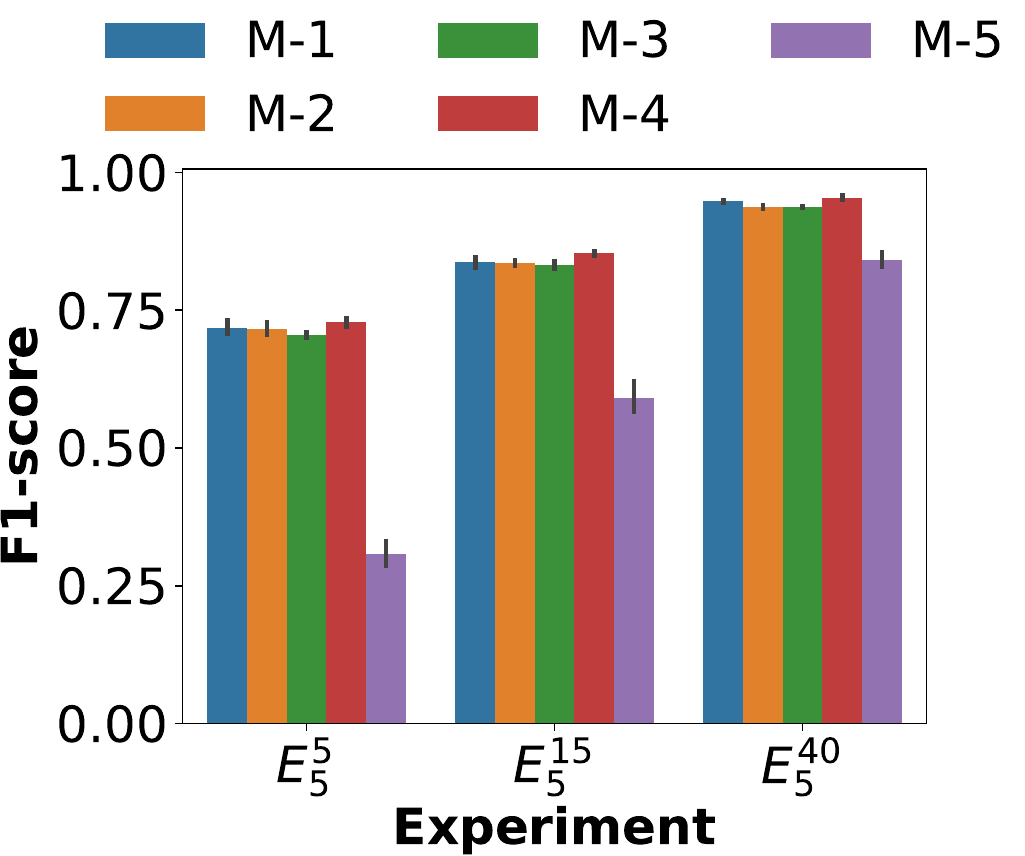}
    \caption{}
    \label{gg2}
\end{subfigure}

\caption{(a) The average running time in log scale (s) and (b) the average maximum F1-score of reconstructing the synthetic networks via five methods for the WPG SICIN.} 
\label{fig: F1time}
\end{figure}

To demonstrate the computational performance of the three incorporated optimization techniques, we devise five methods by combining different techniques and the IP. The details of the five methods are outlined in Table \ref{tab: method_comb}, where the $\checkmark$ denotes that the corresponding method uses the corresponding technique and \textit{Validation} refers to validating the proposed graph via Theorem \ref{theorem1}. We implement each method $10$ times (i.e., $10$ replicates) to reconstruct the synthetic networks and report their average (reconstruction) running time and their average maximum F1-score in Fig. \ref{fig: F1time}. Given that the $95\%$ confidence interval (CI)-based error bars in Fig. \ref{fig: F1time} are small, we are confident about the average values. In each run, we propose $3,000$ samples and use the first $2,000$ samples as a warm-up. The numerical simulations in Fig. \ref{fig: F1time} have been performed on a Windows 10 laptop with a 1.9 GHz Intel Core i7-8650 U CPU and 16 GB RAM.

In Fig. \ref{gg1}, M-1 and M-2 have the shortest running time and the difference in running time between M-1, M-2, and the rest becomes larger as the amount of data increases from $E_5^{5}$ to $E_5^{40}$, indicating that traversing edge lists rather than adjacency matrix to calculate the likelihood significantly reduces the computational load and this reduction is significant when more failure data is used. The comparable running time between M-1 and M-2 indicates that applying graph validation according to Theorem \ref{theorem1} brings \textit{almost no} improvement in computational efficiency. This is because proposing networks by removing edges in the TNT sampler can easily generate disconnected node pairs, and once finding such node pairs, constraints \ref{constraint1}--\ref{constraint6} are broken, leading to saving the time that would be otherwise needed to check other node pairs. Comparing M-3 and M-4, replacing the random sampler with the TNT sampler reduces the computational load because the TNT sampler proposes more valid topological variation by removing edges rather than adding invalid edges that would be later rejected. The running time of M-4 is greater than that of M-5 in $E_5^{5}$ while it is vice versa in $E_5^{15}, E_5^{40}$. Given the smaller amount of cascading failure data in $E_5^{5}$, the proposed networks can easily be accepted without applying the IP to check any topological constraint in advance. Thus, the sampling process is completed in a shorter time in M-5, where no extra operations on checking the network topology are required. However, for larger cascading failure data in $E_5^{15}$ and $E_5^{40}$, and without verifying the network topology for topological constraints \ref{constraint1}--\ref{constraint10} in advance, the topology candidates proposed by the random sampler are more likely to be rejected as they are not supported by the cascading failure data (low likelihood). Then, the time spent on proposing such invalid samples is compensated by the time for checking the topological constraints, leading to shorter times in M-4 than M-5 for $E_5^{15}$ and $E_5^{40}$.

In Fig. \ref{gg2}, the first four methods lead to roughly the same and better average maximum F1-score than M-5, which indicates the power of the IPs' incorporation. Nevertheless, M-4, which applies the random sampler, is slightly better than the previous three methods that use the TNT sampler, as some edges that are present in the target network have already been removed by the TNT sampler in M-1, M-2, and M-3, and the random sampler has a higher chance of adding edges than the TNT sampler, thus the former is more likely to recover those edges that should exist in the target network but have already been removed. The average running times and the average maximum F1-scores of different experiments are provided in Table \ref{table: timecompare}. Considering the trade-off between the running time and the reconstruction accuracy, M-1 is identified as the best method.

\begin{table}[t]
\caption{The average running time (s) and the average maximum F1-score of five methods for the WPG SICIN. The bold font denotes the top two methods.}\label{table: timecompare}
\centering
\begin{tabular}{|c|c|c|c|}
\hline
\diagbox{\textrm{Method}}{Scenario}&
\textrm{$E_5^{5}$}&
\textrm{$E_5^{15}$}&
\textrm{$E_5^{40}$}
\\
\hline
\multicolumn{4}{|c|}{Average running time (s)} \\
\hline
\textbf{M-1} & \textbf{57.608} & \textbf{162.836} & \textbf{712.096}\\
\hline
\textbf{M-2} & \textbf{54.257} & \textbf{161.850} & \textbf{702.410}\\
\hline
M-3 & 95.876 & 368.205 & 2067.501\\
\hline
M-4 & 261.652 & 697.319 & 3762.572 \\
\hline
M-5 & 150.339 & 944.409 & 20305.656 \\
\hline
\multicolumn{4}{|c|}{Average maximum F1-score} \\
\hline
\textbf{M-1} & \textbf{0.718} & \textbf{0.837} & \textbf{0.948}\\
\hline
M-2 & 0.716 & 0.836 & 0.938\\
\hline
M-3 & 0.705 & 0.832 & 0.937\\
\hline
\textbf{M-4} & \textbf{0.728} & \textbf{0.853} & \textbf{0.954} \\
\hline
M-5 & 0.307 & 0.591 & 0.841 \\
\hline
\end{tabular}
\end{table}

Table \ref{table: timecompare} indicates that M-4 achieves the best average maximum F1-score, which stems from the choice of the random sampler. This motivates us to further investigate the effect of sampler choice on the average running time and the average maximum F1-score. To that end, we conduct similar simulations for three methods (M-1, M-1-r, and M-2-r) along with $2$ experiments ($E_5^5$ and $E_5^{15}$) where the modified methods M-1-r and M-2-r are simply obtained from the methods M-1 and M-2 by only replacing the TNT sampler with the random sampler, respectively (i.e., keeping the rest of the techniques unchanged in the configuration). The average running times and the average maximum F1-scores of different experiments are provided in Table \ref{table: F1timecompare}. Considering the trade-off between the running time and the reconstruction accuracy, M-2-r (i.e., Configuration: IP + Random + Edge List) and M-1 (i.e., Configuration: IP + TNT + Edge List + Validation) are suggested as the best method among the three methods for the experiments $E_5^5$ and $E_5^{15}$, respectively. It is worth noting that in the case of $5$ cascading failure scenarios, the best method is no longer identified as M-1. However, increasing the amount of data to the $15$ cascading failure scenarios, the best method is still identified as M-1. Such an observation highlights the positive impact of collecting more cascading failure scenarios on the reconstruction accuracy. Furthermore, it is remarkable that depending on the considered weighted sum of effects of the average (reconstruction) running time and the average maximum F1-score, e.g., $w/T_{\text{avg}} + (1-w) F_1$ (weight: $w \in [0,1]$) and the choice of weight $w$, the best method could be identified differently. Therefore, the human operator can play a significant role in terms of striking a balance between the average (reconstruction) running time and the average maximum F1-score via properly selecting the weighted sum coefficient $w$ based on the physics-informed priorities.

\begin{table}[H]
\caption{The average running time (s) and the average maximum F1-score of three methods for the WPG SICIN. The bold font denotes the top two methods. The Italic font denotes the partially top two methods.}\label{table: F1timecompare}
\centering
\begin{tabular}{|c|c|c|}
\hline
\diagbox{\textrm{Method}}{Scenario}&
\textrm{$E_5^{5}$}&
\textrm{$E_5^{15}$}
\\
\hline
\multicolumn{3}{|c|}{Average running time (s)} \\
\hline
\textbf{M-1} & \textbf{30.553} & \textbf{100.247} \\
\hline
M-1-r & 115.430 & 278.120\\
\hline
\textbf{M-2-r} & \textbf{108.405} & \textbf{262.458} \\
\hline
\multicolumn{3}{|c|}{Average maximum F1-score} \\
\hline
\textit{M-1} & 0.714 & \textbf{0.853} \\
\hline
\textbf{M-1-r} & \textbf{0.741} & \textbf{0.856} \\
\hline
\textit{M-2-r} & \textbf{0.739} & 0.844 \\
\hline
\end{tabular}
\end{table}

\section{Conclusion}\label{sec: conclusions}

This paper presents a Bayesian approach for topology reconstruction of the interdependent critical infrastructure (ICI) networks based on observed cascading failure data, which is the first application of topology reconstruction methods for ICI networks. The hierarchical stochastic block model (HSBM) is used to capture the clustering and hierarchical structure of ICI networks. The computational complexity resulting from the exponential growth of the potential topology is addressed with a new infrastructure-dependent proposal (IP), taking into account the topological constraints of ICI networks. We show the superiority of the IP-based method over a graph-embedding-based approach, namely variational graph auto-encoder (VGAE), in terms of the accuracy of the network reconstruction. Furthermore, we propose three optimization techniques to significantly improve the sampling efficiency. Results of a synthetic network case study demonstrate the effectiveness and transferability of the proposed approach, making it an appropriate choice for the topology reconstruction of the real-world ICI networks, wherein data on the topology is often unavailable either due to security concerns or decentralized operations of different infrastructure sectors. The more accurately we reconstruct the topology of real-world ICI networks, the greater the potential for improved risk mitigation and economic benefits, as it provides policymakers and practitioners with a clearer basis for informed decision-making. We further verify the applicability of the IP-based method to larger ICI networks using a $120 \times 120$ random WPG ICI network where the edge probability $p$ is considered to be equal to $0.5$ for feasible edges, ensuring the satisfaction of topological constraints \ref{constraint1}--\ref{constraint10} and highlighting the fact that for relatively dense (due to $p = 0.5$) and large-scale ICI networks, increasing the number of cascading failure scenarios and the failure length at the expense of higher computational cost could lead to an improvement in the achieved fair/moderate network reconstruction accuracy (i.e., F1-score = $0.664$). Finally, we conduct sensitivity analyses of the network reconstruction accuracy with respect to the two factors: \textit{(i)} the ratio of the initial failed nodes, and \textit{(ii)} the noise level in the cascading failure data. The former sensitivity analysis suggests that the moderate-value choices of the initial failed nodes lead to better network reconstruction accuracy. The latter suggests that the noise in the cascading failure data has an adverse effect on the network reconstruction accuracy. Furthermore, both sensitivity analyses corroborate that the IP-based network reconstruction approach outperforms the non-IP-based counterpart by achieving a less sensitive network reconstruction (in terms of accuracy) against the two factors.

 Future work can be carried out to thoroughly investigate the influence of cascading failure data, network structure, and network size on the reconstruction accuracy. While the proposed approach uses the susceptible–infected (SI) epidemic model to simulate cascading failures in infrastructures, future work can explore more advanced types of failure simulation models (and their stochastic variants) such as susceptible–infected–recovered (SIR). As another future direction, one can think of proposing a generalized probabilistic approach to evaluate infrastructure interdependencies in the case of \textit{uncertain} and \textit{time-varying} ICI networks.

\section*{acknowledgment}

This material is based upon work supported by the National Science Foundation under Grant 1944559.

\section*{Data Availability}

The data and software that support the findings of this article are openly available in Github at \href{https://nam04.safelinks.protection.outlook.com/?url=https\%3A\%2F\%2Fgithub.com\%2FMirSaleh\%2FPRE-Paper-Source-Code&data=05\%7C02\%7Chiba.baroud\%40vanderbilt.edu\%7C71c1cbf020064e59b42c08ded5f8fa23\%7Cba5a7f39e3be4ab3b45067fa80faecad\%7C0\%7C0\%7C639183462480027310\%7CUnknown\%7CTWFpbGZsb3d8eyJFbXB0eU1hcGkiOnRydWUsIlYiOiIwLjAuMDAwMCIsIlAiOiJXaW4zMiIsIkFOIjoiTWFpbCIsIldUIjoyfQ\%3D\%3D\%7C0\%7C\%7C\%7C&sdata=XtjYp4\%2FpBfAyRhHPUhjgeG0\%2FfNv2sIOPPWkyllXIFZ4\%3D&reserved=0}{https://github.com/MirSaleh/PRE-Paper-Source-Code}.

\appendix

\section{Space of Candidate Networks} \label{app: reducespace}

Since the effect of reduction on the sampling space of network topology by considering additional constraints is universal among different networks, we only select one network $I\in\mathcal{I}$ and calculate the total number of potential topologies with and without topological constraints \ref{constraint1}--\ref{constraint10} to demonstrate the significant decrease in the sampling space.

Before imposing additional topological constraints, the number of potential edges in the candidate network $I$ is upper bounded by $\binom{|{{\mathcal V}^I}|}{2}$. Since every edge in the above $\binom{|{{\mathcal V}^I}|}{2}$ edges has two states (existent or non-existent), the total number of the candidate topology is $2^{\binom{|{{\mathcal V}^I}|}{2}}$ whose complexity is an exponential of a quadratic term in terms of $|{{\mathcal V}^I}|$.

After imposing topological constraints \ref{constraint1}--\ref{constraint10}, the total number of potential topologies for the network $I$ is equal to the total number of bipartite graphs consisting of $|{{\mathcal V}^I}| = |\mathcal{V}_\text{s}^I| + |\mathcal{V}_\text{d}^I|$ non-isolated vertices (with $|\mathcal{V}_\text{s}^I|$ vertices and $|\mathcal{V}_\text{d}^I|$ vertices in the supply part and the demand part, respectively) that can be computed as follows:
\begin{align}
& \sum_{k=0}^{|\mathcal{V}_\text{d}^I|}~(-1)^k \binom{|\mathcal{V}_\text{d}^I|}{k}(2^{|\mathcal{V}_\text{d}^I|-k}-1)^{|\mathcal{V}_\text{s}^I|}. \label{eq:noisolated}
\end{align}
\begin{proof}
To derive \eqref{eq:noisolated}, we count the total number of the relations $r: \{1,\dots,|\mathcal{V}_\text{d}^I|\} \to \{1,\dots,|\mathcal{V}_\text{s}^I|\}$ for which the following conditions hold:
\begin{subequations} \label{totalnum}
\begin{align}
    & r^{-1}(j) \ne \emptyset,~\forall j \in \{1,\dots,|\mathcal{V}_\text{s}^I|\} \label{tot1}\\
    & \bigcup_{j \in \{1,\dots,|\mathcal{V}_\text{s}^I|\}}~r^{-1}(j) = \{1,\dots,|\mathcal{V}_\text{d}^I|\} \label{tot2}
\end{align}    
\end{subequations}where $r^{-1}(j) := \{i \in \{1,\dots,|\mathcal{V}_\text{d}^I|\}: (i,j) \in r\}$. Observe that the satisfaction of \eqref{tot1} is equivalent to the non-isolatedness of all the supply vertices in the underlying graph, i.e., $\{1,\dots,|\mathcal{V}_\text{s}^I|\}$. Similarly, the satisfaction of \eqref{tot2} is equivalent to the non-isolatedness of all the demand vertices in the underlying graph, i.e., $\{1,\dots,|\mathcal{V}_\text{d}^I|\}$. 

For all $i \in \{1,\dots,|\mathcal{V}_\text{d}^I|\}$, let us define the following sets:
\begin{subequations} \label{Xsets}
\begin{align}
    & X_i := \{r : \eqref{tot1}~\text{holds},~i \notin \bigcup_{j \in \{1,\dots,|\mathcal{V}_\text{s}^I|\}}~r^{-1}(j)\} \label{Xsets1}\\
    & \bar{X}_i := \{r \in S: r \notin X_i\},~S := \{r : \eqref{tot1}~\text{holds}\}. \label{Xsets2}
\end{align}    
\end{subequations}Note that according to \eqref{Xsets}, the set $\bigcap_{i \in \{1,\dots,|\mathcal{V}_\text{d}^I|\}}~\bar{X}_i$ specifies all the relations satisfying the conditions in \eqref{totalnum}. Observe that $|S| = (-1)^k \binom{|\mathcal{V}_\text{d}^I|}{k} (2^{|\mathcal{V}_\text{d}^I|-k}-1)^{|\mathcal{V}_\text{s}^I|}$ holds for $S$ in \eqref{Xsets2} and $k = 0$. According to \eqref{Xsets1}, we also observe that $|\bigcap_{h \in \{1,\dots,k\}}~X_{i_h}| = (2^{|\mathcal{V}_\text{d}^I|-k}-1)^{|\mathcal{V}_\text{s}^I|}$ hold for all $k \in \{1,\dots,|\mathcal{V}_\text{d}^I|\}$ and $\binom{|\mathcal{V}_\text{d}^I|}{k}$ possible combination $\{i_1,\dots,i_k\}$ can be chosen from $\{1,\dots,|\mathcal{V}_\text{d}^I|\}$. Note that the non-emptiness property in \eqref{tot1} is linked to the corresponding cardinality $2^{|\mathcal{V}_\text{d}^I|-k}-1$ and the exponent ${|\mathcal{V}_\text{s}^I|}$ (i.e., the number of multiplications) is linked to the number of $r^{-1}(j)$ for all $j \in \{1,\dots,{|\mathcal{V}_\text{s}^I|}\}$. Also, note that the property in \eqref{tot2} is linked to the equation $\bigcap_{i \in \{1,\dots,|\mathcal{V}_\text{d}^I|\}}~\bar{X}_i = \{1,\dots,|\mathcal{V}_\text{d}^I|\}$.

Thus, applying the complementary form of the inclusion-exclusion principle, we get \eqref{eq:noisolated}.
\end{proof}

It is worth noting that following a similar path (exchanging the supply nodes and the demand nodes), an alternative formula for \eqref{eq:noisolated} can be obtained as follows:
\begin{align*}
& \sum_{k=0}^{|\mathcal{V}_\text{s}^I|}~(-1)^k \binom{|\mathcal{V}_\text{s}^I|}{k}(2^{|\mathcal{V}_\text{s}^I|-k}-1)^{|\mathcal{V}_\text{d}^I|}.
\end{align*}

Considering a network with $2$ supply nodes on the top level and $3$ demand nodes on the bottom level, the total number of its potential topologies is $25$ according to \eqref{eq:noisolated}, which is a significant decrease from the original $2^{\binom{5}{2}} = 1024$ different topology candidates.

\section{Convergence Proof of Alg. $1$} \label{app:converge}

The proof of Alg. $1$ can be detailed as follows:
\begin{proof}
The limit theorem of Markov Chains states that \textit{if the Markov chain is \textbf{Irreducible} and \textbf{Aperiodic}, the chain will converge to a unique stationary distribution}. Therefore, to prove the convergence of the Markov Chain in Alg. $1$, we need to prove the above two properties for this Markov Chain. Let $\bm{G}_0$, $\bm{G}_1$, \dots, be the Markov chain $\mathcal{G}$ of Alg. $1$ where $\bm{G}_i \in \mathcal{G}$ is the $i$-th state of the ICI networks and the movement from state $\bm{G}_i$ to state $\bm{G}_{i+1}$ is strictly governed by the infrastructure-dependent proposal as shown in Fig. \ref{fig: mcmc}.

\textbf{Irreducible}. A Markov chain is irreducible if every state can be reached from every other state, which can mathematically be formalized as $\forall \bm{G}_i, \bm{G}_j \in \mathcal{G}, P(\bm{G}_j|\bm{G}_i) > 0$. To prove the irreducibility of $\mathcal{G}$, we need to show that we can go from any state $\bm{G}_i \in \mathcal{G}$ following the infrastructure-dependent proposal to any other state $\bm{G}_j \in \mathcal{G}$ in finite steps. Denoting the edge sets of these two states as $\mathcal{E}^{\bm{G}_i}, \mathcal{E}^{\bm{G}_j}$, their shared edge set as $\mathcal{E}^{\bm{G}_i}\cap\mathcal{E}^{\bm{G}_j}$, and their difference as $\mathcal{E}^{\bm{G}_i-\bm{G}_j} = \mathcal{E}^{\bm{G}_i} - \mathcal{E}^{\bm{G}_j}, \mathcal{E}^{\bm{G}_j-\bm{G}_i} = \mathcal{E}^{\mathbf{G}_j} - \mathcal{E}^{\bm{G}_i}$. We first construct a medium graph $\bm{G}_{k}$ by adding all edges in $\mathcal{E}^{\bm{G}_j - \bm{G}_i}$ consecutively to the graph $\bm{G}_{i}$. Since $\mathcal{E}^{\bm{G}_j - \bm{G}_i}\subset \mathcal{E}^{\bm{G}_j}$ and $\mathcal{E}^{\bm{G}_j}$ satisfies constraints \ref{constraint7}--\ref{constraint8}
, all edges in $\mathcal{E}^{\bm{G}_j - \bm{G}_i}$ point from high-level nodes to low-level nodes in the feasible set $\mathcal{S}$ and thus adding these edges during graph construction causes no cycles in $\bm{G}_k$, i.e., no violation against constraints \ref{constraint7}--\ref{constraint10}
. Besides, adding edges trivially satisfies constraints \ref{constraint1}--\ref{constraint6}
. Therefore, the graph construction process from $\bm{G}_i$ to $\bm{G}_k$ is strictly governed by the infrastructure-dependent proposal. Then, we remove all edges in $\mathcal{E}^{\bm{G}_i-\bm{G}_j}$ consecutively from the graph $\bm{G}_{k}$ to obtain graph $\bm{G}_j$. Since $\mathcal{E}^{\bm{G}_j}\subset\mathcal{E}^{\bm{G}_k}$ and $\mathcal{V}^{\bm{G}_j} = \mathcal{V}^{\bm{G}_k}$, the medium graph $\bm{G}_k$ is formed by adding extra edges $\mathcal{E}^{\bm{G}_i-\bm{G}_j}$ to the target graph $\bm{G}_j$, which is also a spanning tree of $\bm{G}_k$. Because $\mathcal{E}^{\bm{G}_i-\bm{G}_j}\cap\mathcal{E}^{\bm{G}_j} = \varnothing$, removing edges in $\mathcal{E}^{\bm{G}_i-\bm{G}_j}$ will not delete any edge in the spanning three $\bm{G}_j$ of $\bm{G}_k$ and this well-preserved spanning three $\bm{G}_j$ guarantees that the medium graph $\bm{G}_{k}$ always satisfies topological constraints \ref{constraint1}--\ref{constraint10} during edge removal. Therefore, the edge removal process from $\bm{G}_k$ to $\bm{G}_j$ is also strictly governed by the infrastructure-dependent proposal. Concatenating the process of edge addition and edge removal gives a way to go from $\bm{G}_i$ to $\bm{G}_j$ under the supervision of the infrastructure-dependent proposal as Fig. \ref{fig: add-remove}, thus the MC $\mathcal{G}$ is irreducible.

\begin{figure}[t]
\centering
\includegraphics[width=0.9\linewidth]{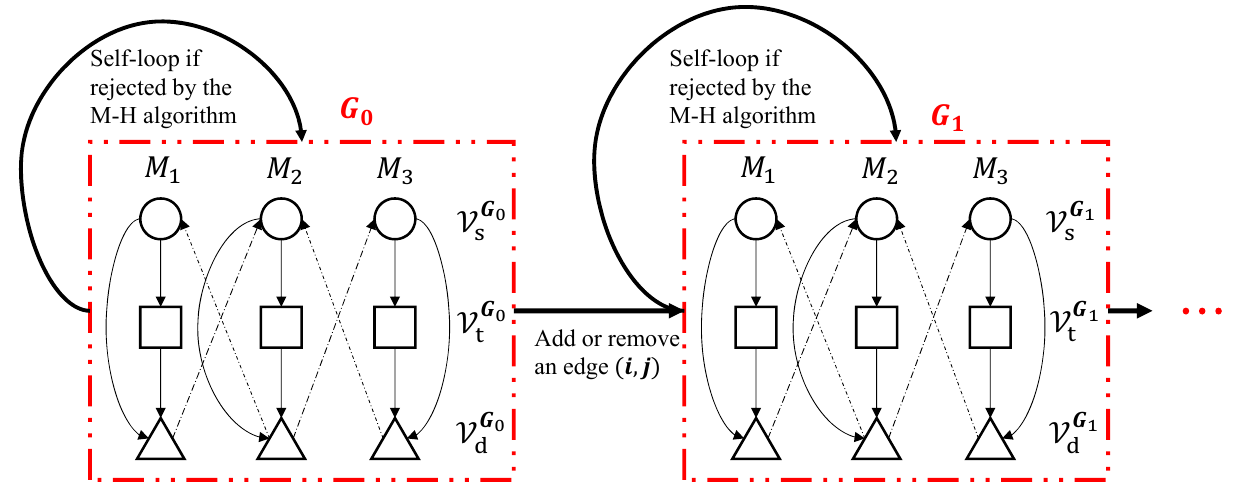}
\caption{The Markov Chain $\mathcal{G}$ of Alg. $1$.} 
\label{fig: mcmc}
\end{figure}

\begin{figure}[t]
\centering
\includegraphics[width=0.9\linewidth]{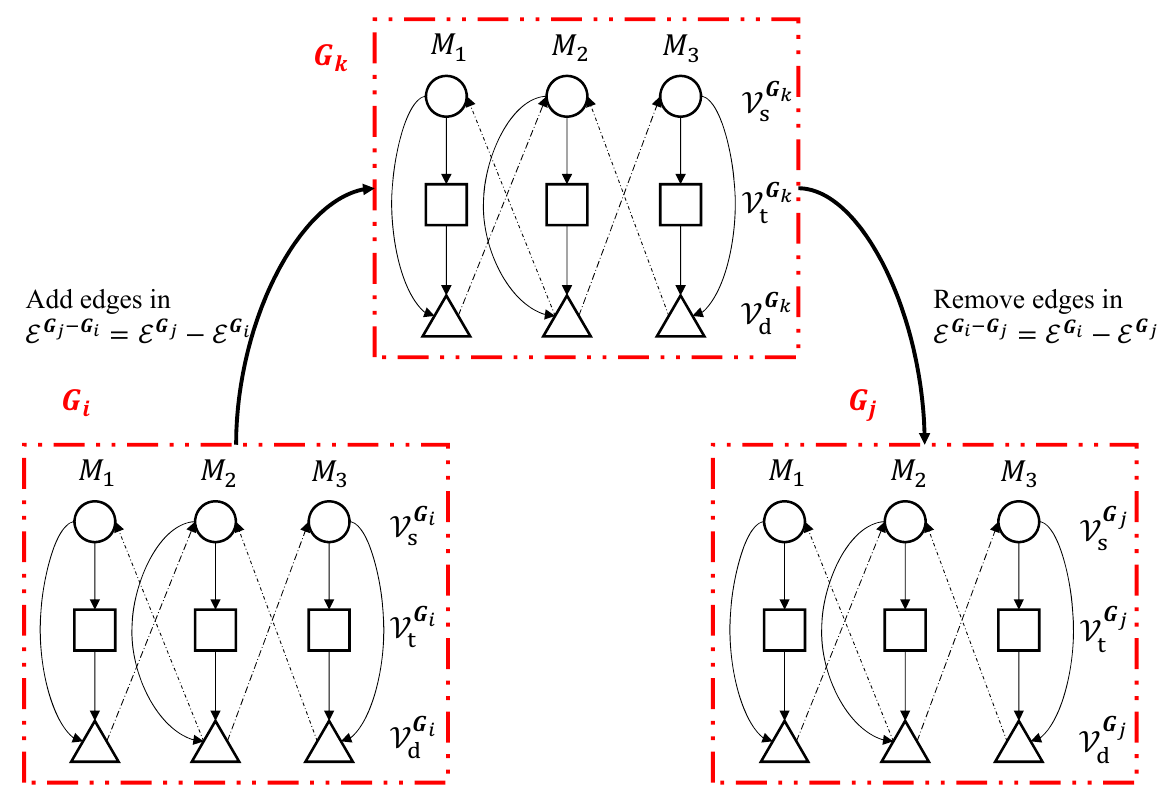}
\caption{The irreducibility of the Markov Chain $\mathcal{G}$.} 
\label{fig: add-remove}
\end{figure}

\textbf{Aperiodic}. A Markov chain is aperiodic if every state is aperiodic. One easy way to determine whether a state $i$ is aperiodic is to check whether state $i$ has a self-loop, i.e., whether MC remains at the same state after moving one step forward from that state. Since every proposal in the M-H algorithm is rejected probabilistically, every state $\bm{G}_i \in \mathcal{G}$ has a self-loop and therefore $\mathcal{G}$ is aperiodic.

Combining the irreducible and aperiodic properties, we conclude that $\mathcal{G}$ will converge to a unique stationary distribution.
\end{proof}

\section{Proof of Theorem 1} \label{sec:theorem1}

First, we propose Lemma \ref{lemma} that will be used to prove Theorem \ref{theorem1} in Section \ref{sec: computationopt}.

\begin{lemma} \label{lemma}
In multi-layer networks $\mathcal{M}$, $\mathcal{I}$, constraints \ref{constraint9}--\ref{constraint10} hold automatically given that constraints \ref{constraint7}--\ref{constraint8}
are satisfied.
\end{lemma}
\begin{proof}
Assuming that we have a multiplex network $\bm{G} = \{\mathcal{M}, \mathcal{I}\}$ that satisfies constraints \ref{constraint7}--\ref{constraint8}, then edges in the edge set $\mathcal{E}^{\bm{G}}$ are all from the feasible set $\mathcal{S}$, which only contains edges from the high-level nodes to low-level nodes for $\mathcal{M}$ and edges from the low-level nodes to high-level nodes for $\mathcal{I}$. Therefore, no backward edges are allowed, and thus edges in the feasible set cannot be used to close a path and form a cycle in $\bm{G}$, i.e., constraints \ref{constraint9}--\ref{constraint10} hold.
\end{proof}

\begin{proof}
Based on how we obtain the current proposal $\bm{A}'$ from the accepted proposal $\bm{A}$ in the last iteration of the M-H algorithm, we divide Theorem \ref{theorem1} into two cases: adding an edge $(i,j)$ or removing an edge $(i,j)$, and prove it in these two cases, respectively.

When an edge $(i,j)$ is added to get $\bm{A}'$ from $\bm{A}$, the path set $\mathcal{P}' = \mathcal{P}\cup\mathcal{P}^{(i,j)}$ contains the path set $\mathcal{P}$ in the original proposal $\bm{A}$ and the extra path set $\mathcal{P}^{(i,j)}$ caused by adding the edge $(i, j)$. By the given condition in Theorem \ref{theorem1} that the set of paths $\mathcal{P}$ in $\bm{A}$ has already satisfied the constraints \ref{constraint1}--\ref{constraint6}, the constraints \ref{constraint1}--\ref{constraint6} are trivially satisfied by $\mathcal{P}'$ in $\bm{A}'$ since we can always find a path $i \rightarrow j \in \mathcal{P}\subset \mathcal{P}'$ to connect two nodes required by constraints \ref{constraint1}--\ref{constraint6}. Besides, edges in the edge set $\mathcal{E}$ and the current added edge $(i,j)$ are all chosen from the feasible set $\mathcal{S}$ that satisfies constraints \ref{constraint7}--\ref{constraint8}, so the current edge set $\mathcal{E}' = \mathcal{E}\cup\{(i,j)\}$ also satisfies constraints \ref{constraint7}--\ref{constraint8}, which leads to the satisfaction of constraints \ref{constraint9}--\ref{constraint10} by Lemma \ref{lemma}.

When an edge $(i,j)$ is removed to get $\bm{A}'$ from $\bm{A}$, all edges in the current graph $\bm{G}'$ are from the $\bm{G}$, i.e., $\mathcal{E}' = \mathcal{E}\backslash(i,j)\subset \mathcal{E}$. By Theorem \ref{theorem1}, graph $\bm{G}$ has already satisfied constraints \ref{constraint7}--\ref{constraint8}, i.e., all edges in $\mathcal{E}$ are from the feasible set $\mathcal{S}$. Therefore, all edges in $\mathcal{E}'$ also satisfy constraints \ref{constraint7}--\ref{constraint8}, and subsequently satisfy constraints \ref{constraint9}--\ref{constraint10} according to Lemma \ref{lemma}. Proof of $\bm{A}'$ satisfying the other six constraints \ref{constraint1}--\ref{constraint6} is similar, so we only use the first constraint, i.e., constraint \ref{constraint1} as an example, and the same idea can be applied in proving the remaining constraints, i.e., constraints \ref{constraint2}--\ref{constraint6}.

To prove that $\bm{A}'$ satisfies constraint \ref{constraint1}, $\forall u\in \mathcal{V}_{\text{s}}^{\bm{G}'}$, we need to find a corresponding path in $\bm{G}'$ starting from $u$ and ending at $v\in\mathcal{V}_{\text{d}}^{\bm{G}'}$. Since $\bm{G}$ has already satisfied constraint \ref{constraint1}, we can find a path $\bm{P}^{\bm{G}}$ starting from $u$ and ending at $v$ in $\bm{G}$. Since edges are only allowed to point from high-level nodes to low-level nodes, the path $\bm{P}^{\bm{G}}$ is either of length $1$, i.e., an edge $uv$ directly from node $u$ to $v$, or of length $2$, i.e., an edge $uw$ followed by another edge $wv$. When $\bm{P}^{\bm{G}}$ is simply an edge $(u, v)$ and if the removed edge $(i,j)$ is not the edge $(u,v)$, i.e., $i \ne u$, $j\ne v$, we can find a path that is the same as $\bm{P}^{\bm{G}}$ in $\bm{G}'$ and constraint \ref{constraint1} holds trivially. If the removed edge $(i,j)$ is the edge $(u,v)$ that constitutes the whole path $\bm{P}^{\bm{G}}$ as shown in Fig. \ref{fig11}, then after removing the edge $(i,j)$, $u$ points to another node $r\in\mathcal{V}_{\text{t}}^{\bm{G}'}\cup\mathcal{V}_{\text{d}}^{\bm{G}'}$ since $\overrightarrow{\bm{A}'}_u\ne\emptyset$ by the given condition in Theorem \ref{theorem1}. If $r\in\mathcal{V}^{\bm{G}'}_{\text{t}}$, we can find another node $z\in\mathcal{V}^{\bm{G}'}_{\text{d}}$ (since we can find another node $z\in\mathcal{V}^{\bm{G}}_{\text{d}}$ as $\bm{G}$ satisfies constraint \ref{constraint4}) towards which node $r$ points, then we find the path $u-r-z$ in $\bm{G}'$ that satisfies constraint \ref{constraint1}. If $r \in \mathcal{V}^{\bm{G}'}_{\text{d}}$, then we also find the path $u-r$ in $\bm{G}'$ that satisfies constraint \ref{constraint1}.

\begin{figure*}
\centering

\begin{subfigure}[b]{0.3\textwidth}
    \centering
    \includegraphics[width=0.9\textwidth]{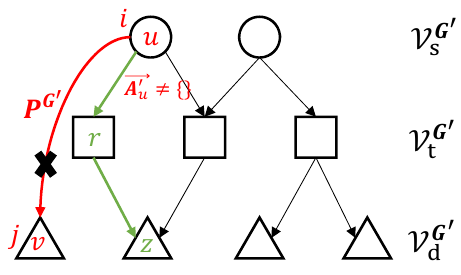}
    \caption{}
    \label{fig11}
\end{subfigure}
\begin{subfigure}[b]{0.3\textwidth}
    \centering
    \includegraphics[width=0.9\textwidth]{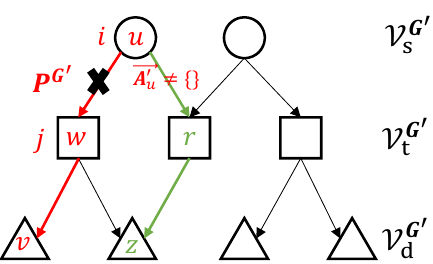}
    \caption{}
    \label{fig22}
\end{subfigure}
\begin{subfigure}[b]{0.3\textwidth}
    \centering
    \includegraphics[width=0.9\textwidth]{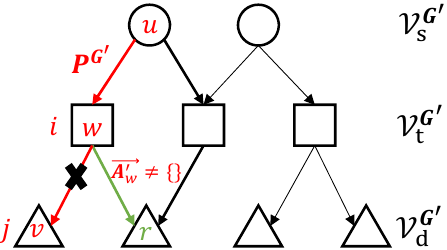}
    \caption{}
    \label{fig33}
\end{subfigure}

\caption{Proof of constraint \ref{constraint1} for (a) the first case, (b) the second case, and (c) the third case.} 
\label{fig: proof1}
\end{figure*}

When the path $\bm{P}^{\bm{G}}$ consists of two edges $u-w-v$, we ignore the analysis of the trivial situation when both of the edges $(u,w)$, $(w,v)$ are not removed but directly focus on situations where either $(u,w)$ or $(w,v)$ is removed as shown by Fig. \ref{fig22}, \ref{fig33}.

When $(u, w)$ is removed as shown in Fig. \ref{fig22}, then $u$ points to another node $r\in\mathcal{V}_{\text{t}}^{\bm{G}'}\cup\mathcal{V}_{\text{d}}^{\bm{G}'}$ since $\overrightarrow{\bm{A}}'_u\ne\emptyset$. If $r\in\mathcal{V}^{\bm{G}'}_{\text{t}}$, we can find another node $z\in\mathcal{V}^{\bm{G}'}_{\text{d}}$ (since we can find another node $z\in\mathcal{V}^{\bm{G}}_{\text{d}}$ as $\bm{G}$ satisfies constraint \ref{constraint4}) towards which node $r$ points, then we find the path $u-r-z$ in $\bm{G}'$ that satisfies constraint \ref{constraint1}. If $r \in \mathcal{V}^{\bm{G}'}_{\text{d}}$, then we also find the path $u-r$ in $\bm{G}'$ that satisfies constraint \ref{constraint1}. When $(w, v)$ is removed as Fig. \ref{fig33}, then $w$ points to another node $r\in\mathcal{V}_{\text{d}}^{\bm{G}'}$ since $\overrightarrow{\bm{A}'}_w\ne\emptyset$. We also find the path $u-w-r$ in $\bm{G}'$ that satisfies constraint \ref{constraint1}.

The above explanation only completes the forward direction $(\Rightarrow)$ of the proof (i.e., the if part). The backward direction $(\Leftarrow)$ of the proof (i.e., the only if part) is as follows: When an edge $(i, j)$ is added to $\bm{G}$ to get $\bm{G}'$, then $\overrightarrow{\bm{A}}'_i$ contains at least $j$ and $\overleftarrow{\bm{A}}'_j$ contains at least $i$. Therefore, $\overrightarrow{\bm{A}}'_i \ne \emptyset$, $\overleftarrow{\bm{A}}'_j \ne \emptyset$. When an edge $(i,j)$ is removed from $\bm{G}$ to get $\bm{G}'$, since constraints \ref{constraint1}--\ref{constraint6} still hold in $\bm{G}'$ after the edge removal according to the condition in Theorem \ref{theorem1}, we can find a path $\bm{P}_1$ starting from node $i$ and find another path $\bm{P}_2$ ending at node $j$. Taking the first edge $ir$ in the path $\bm{P}_1$ and the last first edge $zj$ in the path $\bm{P}_2$, then $\overrightarrow{\bm{A}}'_i$ contains at least $r$ and $\overleftarrow{\bm{A}}'_j$ contains at least $z$. Therefore, $\overrightarrow{\bm{A}}'_i \ne \emptyset$, $\overleftarrow{\bm{A}}'_j \ne \emptyset$.
\end{proof}

\bibliography{apsPREarXiv}% Produces the bibliography via BibTeX.

\end{document}